# Reinforcement Learning for Agents with Many Sensors and Actuators Acting in Categorizable Environments

**Josep M Porta**                                                    PORTA@SCIENCE.UVA.NL
*IAS Group, Informatics Institute*
*University of Amsterdam*
*Kruislaan 403, 1098SJ, Amsterdam, The Netherlands*

**Enric Celaya**                                                     CELAYA@IRI.UPC.EDU
*Institut de Robòtica i Informàtica Industrial*
*Spanish Council of Scientific Research (CSIC)*
*Llorens i Artigas 4-6, 08028, Barcelona, Spain*

## Abstract

In this paper, we confront the problem of applying reinforcement learning to agents that perceive the environment through many sensors and that can perform parallel actions using many actuators as is the case in complex autonomous robots. We argue that reinforcement learning can only be successfully applied to this case if strong assumptions are made on the characteristics of the environment in which the learning is performed, so that the relevant sensor readings and motor commands can be readily identified. The introduction of such assumptions leads to strongly-biased learning systems that can eventually lose the generality of traditional reinforcement-learning algorithms.

In this line, we observe that, in realistic situations, the reward received by the robot depends only on a reduced subset of all the executed actions and that only a reduced subset of the sensor inputs (possibly different in each situation and for each action) are relevant to predict the reward. We formalize this property in the so called *categorizability assumption* and we present an algorithm that takes advantage of the categorizability of the environment, allowing a decrease in the learning time with respect to existing reinforcement-learning algorithms. Results of the application of the algorithm to a couple of simulated realistic-robotic problems (landmark-based navigation and the six-legged robot gait generation) are reported to validate our approach and to compare it to existing flat and generalization-based reinforcement-learning approaches.

## 1. Introduction

The division between knowledge-based and behavior-based artificial intelligence has been fundamental to achieving successful applications within the field of autonomous robots (Arkin, 1998). However, up to now, this division has had few repercussions for reinforcement learning. Within artificial intelligence, reinforcement learning has been formalized in a very general way borrowing ideas from the dynamic programming and decision-theory fields. Within this formalization, the objective of reinforcement-learning methods is to establish a correct mapping from a set of abstract observations (formalized as *states*) to a set of high level actions, without being worried about how these sets of states and actions are defined (for an introduction to reinforcement learning you can check Kaelbling, Littman, & Moore, 1996; Sutton & Barto, 1998, among many others). Algorithms developed within this general framework can be used in different fields without any modification. For each particular application, the definition of the sets of states and actions is the responsibility of the programmer and is not supposed to be part of the reinforcement-learning problem. However, as clearly pointed by Brooks (1991), in autonomous robots the major hurdles are those related with perception and action representations. For this reason, in a robotic task, what traditional reinforcement-learning research assumes to be the major problem (connecting states and actions) is simpler than what it assumes as given (the definition of states and actions). The consequence is that existing reinforcement-learning methods are best suited for problems that fall into the symbolic artificial intelligence domain than for those that belong to robotics. Due to the generality of existing reinforcement-learning algorithms, a robotic problem can be analyzed and re-formulated so that it can be tackled with the available reinforcement-learning tools but, in many cases, this re-formulation is too awkward introducing unnecessary complexity in the learning process. The alternative we explore in this paper is a new reinforcement-learning algorithm that can be applied to robotic problems as they are, without any re-formulation.

As Brooks (1991) remarked, dealing with a real environment is not necessarily a problem since real environments have properties that can be exploited to reduce the complexity of the robot's controller. In Brooks' works, we can find simple robot controllers that achieve very good performance in particular environments. This is clearly in contrast with the generality pursued within reinforcement learning. Following an idea parallel to that of Brooks, in this paper, we present a new reinforcement-learning algorithm that takes advantage of a specific environment-related property (that we call *categorizability*) to efficiently learn to achieve a given task. We formalize the categorizability property and we present a representation system (*partial rules*) to exploit this property. A remarkable feature of this representation system is that it allows generalization in both the spaces of sensors and actions, using a uniform mechanism. This ability to generalize in both the state and action spaces is fundamental to successfully apply reinforcement learning to autonomous robots.

This paper is organized as follows. First, in Section 2, we formalize reinforcement learning from the point of view of its use in the field of autonomous robotics and we describe the problems that make flat (and, in most cases, also generalization-based) reinforcement-learning algorithms not adequate for this case. Section 3 presents the *categorizability assumption* which is plausible in most robotics environments. Then, in Section 4, we describe an alternative reinforcement-learning algorithm that exploits the categorizability assumption to circumvent the problems present in existing approaches. In Section 5, we analyze the points of contact between our proposal and already existing work. Next, in Section 6, we present experiments that validate our approach. The experiments are performed in simulations that mimic realistic robotic applications where the categorizability assumption is likely to be valid. Finally, in Section 7, we conclude by analyzing the strengths and weaknesses of the proposed learning system.

Additionally, Appendix A provides a detailed description of the partial-rule learning algorithm introduced in this paper, Appendix B is devoted to an enhancement on this algorithm to make its execution more efficient, and Appendix C summarizes the notation we use throughout the paper.

## 2. Problem Formalization

For simplicity, we assume that the robot perceives its environment through a set of binary feature detectors[1] $FD = \{fd_i \mid i = 1..n_f\}$. A feature detector can be devised as a process that identifies specific combinations of present (and possibly past) sensor readings. The use of feature detectors is very common in robotics. In this field, feature detectors are defined by the programmer attending to the special characteristics of the environment, the robot sensors, and the task to be executed in order to extract potentially useful information (presence of landmarks or obstacles, . . . ) from raw sensor readings.

In a similar way, instead of working directly with the space of actions provided by the robot motors (that define a too low-level way of controlling the robot), it is a common practice to define a set of *elementary actions* $EA = \{ea_i \mid i = 1..n_e\}$. An elementary action is a specific sequence/combination of motor commands defined by the programmer attending to the characteristics of the robot and the task to be achieved. To simplify, we can assume that elementary actions are of the form $(m_i \leftarrow k)$ $(i \in [1..n_m])$ where $m_i$ is a motor and $k$ a value in the range of valid inputs for the motor $m_i$. This framework is quite flexible since a motor $m_i$ can be either one of the physical motors of the robot or a high-level, abstract motor that combines movements of the actual motors. With this formalization, at a given moment, the robot can execute in parallel as many elementary actions as available motors.

A robot controller can be seen as a procedure that executes (combinations of elementary) actions in response to specific situations (i.e., activation of specific feature detectors) with the objective of achieving a given task. Reinforcement-learning approaches automatically define such a controller using the information provided by the reward signal. In the context of reinforcement learning, the controller is called the *policy* of the learner.

The objective of the value-function-based reinforcement-learning algorithms (the most common reinforcement-learning algorithms) is to predict the reward that can be directly or indirectly obtained from the execution of each action (i.e., of each combination of elementary actions) in each possible situation, described as a combination of active and inactive feature detectors. If this prediction is available, the action to be executed in each situation is the one from which maximum reward is expected.

To predict the reward, classic reinforcement-learning algorithms rely on the Markov assumption, that requires a state signal to carry enough information to determine the effects of all actions in a given situation.[2] Additionally, non-generalizing reinforcement-learning algorithms assume that the states of the system must be learned about independently. So, the information gathered about the effects of an action $a$ in a given state $s$, denoted $Q(s, a)$, cannot be safely transferred to similar states or actions. With this assumption, the cost of a reinforcement-learning algorithm in a general problem is

$$\Omega(n_s\, n_a),$$

where $n_s$ is the number of states and $n_a$ is the number of actions. This is because each action has to be tried as least once in each state. Since the state is defined as the observed

---

1. Non-binary feature detectors providing a discrete range of values can be readily binarized.
2. Non-Markovian problems, when confronted, should be converted into Markovian ones. How to do that is out of the scope of this paper, although it is one of the most relevant points to achieve a successful real-world reinforcement-learning application.

combination of feature detectors, we have that the potential number of states is

$$n_s = 2^{n_f},$$

with $n_f$ the number of feature detectors. Consequently, we have that

$$\Omega(n_s \, n_a) = \Omega(2^{n_f} \, n_a),$$

which is exponential in the number of feature detectors. Since the number of feature detectors used in robotic applications tends to be high, non-generalizing reinforcement learning becomes impractical for realistic problems. This is the well known *curse of dimensionality* introduced by Bellman (1957), whose research presaged some of the work in reinforcement learning.

Although the size of the action set ($n_a$) is as important as the size of the state set ($n_s$) in the curse of dimensionality, less attention is paid to actions in the reinforcement-learning literature. However, a robot with many degrees of freedom can execute many elementary actions simultaneously and this makes the cost of the learning algorithms also increase exponentially with the number of motors of the robot ($n_m$).

Suppose we address the same task but with two different sets of feature detectors $FD_1$ and $FD_2$ such that $FD_1 \subset FD_2$. Using a plain reinforcement-learning algorithm, the cost of finding a proper policy would be larger using the larger set of features ($FD_2$). And this is so even if one of the features in $FD_2 - FD_1$ has a stronger correlation with the reward than any of the features in $FD_1$. Non-generalizing reinforcement-learning algorithms are not able to take advantage of this situation, and, even having better input information, their performance decreases. A similar argument can be made for actions in addition to feature detectors.

Generalizing reinforcement-learning algorithms such as those using gradient-descent techniques (Widrow & Hoff, 1960), coarse codings (Hinton, McClelland, & Rumelhart, 1986), radial-basis functions (Poggio & Girosi, 1990), tile coding (Sutton, 1996) or decision trees (Chapman & Kaelbling, 1991; McCallum, 1995) can partially palliate this problem since they can deal with large state spaces. However, as we approach complex realistic problems, the number of dimensions of the state-space grows to the point of making the use of some of these generalization techniques impractical and other function approximation techniques must be used (Sutton & Barto, 1998, page 209).

Adding relevant inputs or actions to a task should make this task easier or at least not more difficult. Only methods whose complexity depends on the relevance of the available inputs and actions and not on their number would scale well to real domain problems. Examples of systems fulfilling this property are, for instance, the *Kanerva coding* system presented by Kanerva (1988) and the *random representation* method by Sutton and Whitehead (1993). While those systems rely on large collections of fixed *prototypes* (i.e., combinations of feature detectors) selected at random, our proposal is to search for the appropriate prototypes, but using a strong bias so that the search can be performed in a reasonable time. This strong bias is based on the *categorizability assumption* that is a plausible assumption for the case of autonomous robots, which allows a large speed up in the learning process. Additionally, existing systems do not address the problem of determining the relevance of actions, since they assume the learning agent has a single actuator (that is, obviously, the only relevant one). This simple set up is not adequate for robotics. In our approach (presented below), combinations of both feature detectors and elementary actions are considered using a unified framework.

## 3. The Categorizability Assumption

From our experience developing controllers for autonomous robots, we observe that, in many realistic situations, the reward received by the robot depends only on a reduced subset of all the actions executed by the robot and that most of the sensor inputs are irrelevant to predict that reward. Thus, for example, the value resulting from the action of "grasping the object in front of the robot" will depend on what the object is: the object the robot should bring to the user, an electrified cable, or an unimportant object. However, the result will probably be the same whether or not the robot is moving its cameras while grasping the object, if it is day or night, if the robot is, at the same time, checking the distance to the nearest wall, or if it can see a red light nearby or not (aspects, all of them, that may become important in other circumstances).

If an agent observes and acts in an environment where a reduced fraction of the available inputs and actuators have to be considered at a time, we say that the agent is in a categorizable environment.

Categorizability is not a binary predicate but a graded property. In the completely categorizable case, it would be necessary to pay attention to only one sensor/motor in each situation. On the other extreme of the spectrum, if all motors have to be carefully coordinated to achieve the task and the effect of each action could only be predicted by taking into account the value of all feature detectors, we would say that the environment is not categorizable at all.

Since robots have large collection of sensors providing a heterogeneous collection of inputs and many actuators affecting quite different degrees of freedom, our hypothesis is that, in robotic problems, environments are highly categorizable and, in those cases, an algorithm biased by the categorizability assumption would result advantageous.

## 4. Reinforcement Learning in Categorizable Environments: the Partial Rule Approach

To implement an algorithm able to exploit the potential categorizability of the environment, we need a representation system able to transfer information between similar situations and also between similar actions.

Clustering techniques or successive subdivisions of the state space (as, for instance, that presented by McCallum, 1995) focus on the perception side of the problem and aim at determining the reward that can be expected in a given state $s$ considering only some of the feature detectors perceived in that state. This subset of relevant feature detectors is used to compute the expected reward in this state for any possible action $a$ (the $Q(s, a)$ function). However, with this way of posing the problem the curse of dimensionality problem is not completely avoided since some of the features can be relevant for one action but not for another and this produces an unnecessary (from the point of view of each action) differentiation between equivalent situations, decreasing the learning speed. This problem can be avoided by finding the specific set of relevant feature detectors for each action. In this case, the $Q$ function is computed as $Q(f_s(a), a)$, with a state definition that is function of the action under consideration. This technique is used, for instance, by Mahadevan and Connell (1992). Unfortunately, in the problem we are confronting, this is not enough since, in our case, actions are composed by combinations of elementary actions and we also want to transfer reward information between similar combinations of actions. Therefore, we have to estimate $Q(f_s(a), a)$ only taking into account some of the elementary actions that compose $a$. However, in principle, the relevance of elementary actions is function of the situation (or, equivalently, of the state): a given elementary action can be relevant in some situations but not in others. For this reason, the function to approximate becomes $Q(f_s(a), f_a(s))$ where there is a cross-dependency between the state defined as a function of the action, $f_s(a)$, and the action defined as a function of the state, $f_a(s)$. The proposal we detail next solves this cross-dependency by working in the Cartesian product of the spaces of feature detectors and elementary actions combinations.

To formalize our proposal, we introduce some definitions.

We say that the agent *perceives* (or *observes*) a *partial view of order* $k$, $v(fd_{i_1}, \ldots, fd_{i_k})$, $k \leq n_f$ whenever the predicate $fd_{i_1} \wedge \ldots \wedge fd_{i_k}$ holds.[3] Obviously, many partial views can be perceived at the same time.

At a given moment, the agent executes an action $a$ that issues a different command for each one of the agent's motors $a = \{ea_1, \ldots, ea_{n_m}\}$, with $n_m$ the number of motors.

A *partial command of order* $k$, noted as $c(ea_{i_1}, \ldots, ea_{i_k})$, $k \leq n_m$, is *executed* whenever the elementary actions $\{ea_{i_1}, \ldots, ea_{i_k}\}$ are executed simultaneously. We say that a partial command $c$ and an action $a$ are *in accordance* if $c$ is a subset of $a$. Note that the execution of a given action $a$ supposes the execution of all the partial commands in accordance with it.

A *partial rule* $w$ is defined as a pair $w = (v, c)$, where $v$ is a partial view and $c$ is a partial command. We say that a partial rule $w = (v, c)$ is *active* if $v$ is observed, and that $w$ is *used* whenever the partial view $v$ is perceived and the partial command $c$ is executed. A partial rule covers a sub-area of the Cartesian product of feature detectors and elementary actions and, thus, it defines a situation-action rule that can be used to partially determine the actions of the robot in many situations (all those where the partial view of the rule is active). The order of a partial rule is defined as the sum of the order of the partial view and the order of the partial command that compose the rule.

We associate a quantiy $q_w$ to each partial rule. $q_w$ is an estimation of the value (i.e., the discounted cumulative reward) that can be obtained after executing $c$ when $v$ is observed at time $t$:

$$q_w = \sum_{i=0}^{\infty} \gamma^{t+i} \, r_{t+i},$$

with $r_{t+i}$ the reward received by the learner at time step $t+i$ after rule $w$ is used at time $t$. So, a partial rule can be interpreted as: **if** *partial view $v$ is observed* **then** *the execution of partial command $c$ results in value $q_w$*.

---

3. A partial view can also include negations of feature detectors since the non-detection of a feature can be as relevant as its detection.

The objective of the learning process is that of deriving a set of partial rules and adjusting the corresponding $q_w$ values so that the desired task can be properly achieved.

The apparent drawback of the partial-rule representation is that the number of possible partial rules is much larger than the number of state and action pairs: The number of partial rules that can be defined on a set of $n_f$ binary feature detectors and $n_m$ binary motors is $3^{n_f+n_m}$, while the number of different states and action pairs is "only" $2^{n_f+n_m}$. If arbitrary problems have to be confronted (as is the case in synthetic learning situations), the partial-rule approach could not be useful. However, problems confronted by robots are not arbitrary since, as mentioned, environments present regularities or properties (as categorizability) that can be exploited to reduce the complexity of the controller necessary to achieve a given task.

Using the partial-rule framework, the *categorizability assumption* can be formally defined as:

**Definition 1** *We say that an environment/task is highly categorizable if there exists a set of low-order partial rules that allows us to predict the reward with the same accuracy as if statistics for each possible state-action combination were considered. The lower the order of the rules in the controller the higher the categorizability of the environment/task.*

To the extent the categorizability assumption is fulfilled, the number of partial rules necessary to control the robot becomes much smaller than the number of state-action pairs that can be defined using the same sets of feature detectors and elementary actions in which the partial views and partial commands are based. Additionally, categorizability implies that the rules necessary in the controller are mostly those with lower order and this can be easily exploited to bias the search in the space of partial rules. So, if the environment is categorizable, the use of the partial-rule approach can suppose an important increase in the learning speed and a reduction in the use of memory with respect to traditional non-generalizing reinforcement-learning algorithms.

In the following sections, we describe how it is possible to estimate the effect of an action given a fixed set of partial rules. This evaluation, repeated for all actions, is used to determine the best action to be executed at a given moment. Next, we detail how it is possible to adjust the value predictions of a fixed set of partial rules. Finally, we describe how the categorizability assumption allows us to use an incremental strategy in the generation of new partial rules. This strategy results in faster learning than existing generalizing and non-generalizing reinforcement-learning algorithms. All procedures are described in high-level form to make the explanation more clear. Details of their implementation can be found in Appendix A.

### 4.1 Value Prediction using Partial Rules

In a given situation, many partial views are simultaneously active triggering a subset of the partial rules of the controller $C$. We call this subset the *active* partial rules and we denote it as $C'$. To evaluate a given action $a$ we only take into account the rules in $C'$ with a partial command in accordance with $a$. We denote this subset as $C'(a)$. Note that, in our approach, when we refer to an action, we mean the corresponding set of elementary actions (one per motor) and not a single element, as it is the general case in reinforcement learning.

Every rule $w = (v, c)$ in $C'(a)$ provides a value prediction for $a$: the $q_w$ associated with the partial rule. This is an averaged value that provides no information about the accuracy of this prediction. As also pointed by Wilson (1995), we should favor the use of the partial rules with a high accuracy in value prediction or, as we say it, rules with a *high relevance.*

It seems clear that the relevance of a rule ($\rho_w$) depends on the distribution of values around $q_w$. Distributions with low dispersion are indicative of coherent value predictions and, so, of a highly relevant rule. To measure this dispersion we maintain an error estimation $e_w$ on the approximation of $q_w$. Another factor (not used by Wilson, 1995) to be taken into account in the relevance determination is the confidence on the $q_w$ and $e_w$ statistics: low confidence (i.e., insufficiently sampled) measures of $q_w$ and $e_w$ should reduce the relevance of the rule. The confidence on the value prediction for a given rule ($c_w$) is a number in the interval $[0, 1]$, initialized as 0, and increasing as the partial rule is used (i.e., the rule is active and its partial command is executed). The confidence would only decrease if the value model for a given partial rule is consistently wrong.

Using the confidence, we approximate the real error in the value prediction for a partial rule $w$ as

$$\epsilon_w = e_w \, c_w + \overline{e} \, (1 - c_w),$$

where value $\overline{e}$ is the average error on the value prediction. Observe that the importance of $\overline{e}$ is reduced as the confidence increases and, consequently, $\epsilon_w$ converges to $e_w$.

With the above definitions, the relevance of a partial rule can be defined as

$$\rho_w = \frac{1}{1 + \epsilon_w}.$$

Note that the exact formula for the relevance is not that important as far as $\epsilon_{w1} \leq \epsilon_{w2} \to \rho_{w1} \geq \rho_{w2}$. The above formula provides a value in the range $[0, 1]$ that could be directly used as a scale factor, if necessary.

The problem is then, how can we derive a single value prediction using the $q_w$ statistics of all the rules in $C'(a)$ and its corresponding relevance value, $\rho_w$? Two possible solutions come to mind: using a weighted sum of the values predicted by all these partial rules using the relevance as a weighting factor, or using a competitive approach, in which only the most relevant partial rule is used to determine the predicted value. The weighted sum assumes a linear relation between the inputs (the value prediction provided by each individual rule) and the output (the value prediction for $a$). This assumption has proved powerful in many systems but, in general, it is not compatible with the categorizability assumption since, although each one of the partial rules involved in the sum can be of low order, taking all of them into account means using a large set of different feature detectors and elementary actions to predict the effect of a given action. For this reason, our learning system uses a winner-take-all solution where only the value prediction of the most relevant partial rule is taken into account to predict the value of an action. So, for each action we determine the winner rule

$$w = winner \, (C', a) = \arg \max_{\forall w' \in C'(a)} \{\rho_{w'}\},$$

and we use the range of likely value for this rule, $I_w = [q_w - 2\epsilon_w, q_w + 2\epsilon_w]$, to randomly determine the value prediction for action $a$. The probability distribution inside this interval depends on the distribution we assume for the value.

The procedure just outlined can be used at each time step to obtain a value prediction for each action. The action with the maximal value is the one we want the robot to execute next.

Observe that we obtain a probabilistic value prediction: in the same situation with the same statistics, we can get different value predictions for the same action. In this way, the action that obtains the maximal evaluation is not always the one with maximal $q_w$ and, consequently, we favor the exploration of promising actions. This probabilistic action selection provides an exploratory mechanism that uses more information than typical reinforcement-learning exploration mechanisms (the error and confidence of value predictions is not available in most reinforcement-learning algorithms) and the result is a more sophisticated exploration schema (see Wilson, 1996, for a survey of different exploration mechanisms in reinforcement learning).

## 4.2 Partial Rules Value Adjustment

We adjust the value predictions for *all* the rules in $C'(a)$ where $a$ is the last executed action. For each rule to be adjusted, we have to update its $q_w$, $e_w$, and $c_w$ statistics.

The effect of any action $a$ in accordance with the partial command $c$ attending to a partial rule $w = (v, c)$ can be defined (using a Bellman-like equation) as

$$q_w^* = \overline{r}_w + \gamma \sum_{\forall C'} p(w, C') \, v^*(C'),$$

where $\overline{r}_w$ is the average reward obtained immediately after executing $c$ when $v$ is observed, $\gamma$ is the discount factor used to balance the importance of immediate with respect to delayed reward, $v^*(C')$ represents the goodness (or value) of the situation where rules $C'$ are active, and $p(w, C')$ is the probability of reaching that situation after the execution of $c$ when $v$ is observed. The value of a situation is assessed using the best action executable in that situation

$$v^*(C') = \max_{\forall a'} \{q_w^* | w = winner(C', a')\},$$

since this gives us information about how well the robot can perform (at most) from that situation.

As in many of the existing reinforcement-learning approaches, the values of $q_w$ and $e_w$ for the rules to be adjusted are modified using a temporal difference rule so that they progressively approach $q_w^*$ and the error on this measure. Rules that have a direct relation with the received reward would provide a value prediction ($q_w$) coherent with the actually obtained one and, consequently, after the statistics adjustment, their prediction error will be decreased. Contrariwise, rules not related to the observed reward would predict a value different from the obtained one and their error statistics will be increased. In this way, if a rule is really important for the generation of the received reward, its relevance is increased and if not it is decreased. Rules with low relevance have few chances of being used to drive the robot and, in extreme cases, they could be removed from the controller.

The confidence $c_w$ should also be adjusted. This adjustment depends on how the confidence is measured. If it is only related to the number of samples used in the $q_w$ and $e_w$ statistics, then $c_w$ should be simply slightly incremented every time the statistics of rule $w$ are updated. However, we also decrease the confidence if the value model for a given partial rule is consistently wrong (i.e., the value observed is systematically out of the interval $I_w$).

Observe that our learning rule is equivalent to those used in state-based reinforcement-learning methods. For instance, in Q-learning (Watkins & Dayan, 1992), $Q^*(s, a)$, with $s$ a state and $a$ an action, is defined as

$$Q^*(s, a) = \bar{r}_w + \gamma \sum_{\forall s'} p(s, a, s') \, V^*(s'),$$

with $p(s, a, s')$ the probability of a transition from $s$ to $s'$ when $a$ is executed and

$$V^*(s') = \max_{\forall a'} \{Q^*(s', a')\}$$

In our approach, the set of rules active in a given situation $C'$ plays the role of a state and, thus, $v^*(C')$ and $V^*(s')$ are equivalent. On the other hand, we estimate $q_w^*$ instead of $Q^*(s, a)$, but the rule $w$ includes information about both (partial) state and actions making $q_w^*$ and $Q^*(s, a)$ to play a similar role. The value prediction for a given rule, $q_w$, corresponds to the average of value predictions for the cells of the Cartesian product of feature detectors and elementary actions covered by that rule. In the case of complete rules (i.e., rules involving all the feature detectors and actions for all motors), the sub-area covered by the rule includes only one cell of the Cartesian product and, therefore, if the controller only includes *complete rules*, the just described learning rule is exactly the same as that used in Q-learning. In this particular case, $C'(a)$ is just one rule that, consequently, is the *winner* rule. The statistics for this rule are the same (and are updated in the same way) as those for the $Q(s, a)$ entry of the table used in Q-learning. Thus, our learning rule is a generalization of the learning rule normally used in reinforcement learning.

### 4.3 Controller Initialization and Partial Rule Creation/Elimination

Since we assume we are working in a categorizable environment, we can use an incremental strategy to learn an adequate set of partial rules: we initialize the controller with rules of the lowest order and we generate new partial rules only when necessary (i.e., for cases not correctly categorized using the available set of rules). So, the initial controller can contain, for instance, all the rules of order two that include one feature detector and one elementary action $((v(fd_i), c(ae_j)), (v(\neg fd_i), c(ae_j)) \; \forall i, j)$. In any case, it is sensible to include the empty rule (the rule of order 0, $w_\emptyset$) in the initial controller. This rule is always active and it provides the average value and the average error in the value prediction. Additionally, any knowledge the user has about the task to be achieved can be easily introduced in the initial controller in the form of partial rules. If available, an estimation of the value predictions for the user-defined rules can also be included. If the hand-crafted rules (and their value predictions) are correct the learning process will be accelerated. If they are not correct, the learning algorithm would take care of correcting them.

We create a new rule when a large error in the value prediction is detected. The new rule is defined as a combination of two of the rules in $C'(a)$, that are the rules that forecast the effects of the last executed action, $a$, in the current situation. When selecting a couple of rules to be combined, we favor the selection of those with a value prediction close to the actually observed one, since they are likely to involve features and elementary actions (partially) relevant for the value prediction we try to refine.

The problem is that it is not possible to determine *a priori* whether an incorrectly predicted value would be correctly predicted after some rule adjustments or if it is really necessary to create a new partial rule to account for the received reward. So, if we create new rules when there is a large error in the value prediction, it is possible to create unnecessary rules. The existence of (almost) redundant rules is not necessarily negative, since they provide robustness to the controller, the so called *degeneracy* effect introduced by Edelman (1989). What must be avoided is to generate the same rule twice, since this is not useful at all. Two rules can be identical with respect to lexicographic criteria (they contain the same feature detectors and elementary actions) but also with respect to semantic ones (they get active in the same situations and propose equivalent actions). If identical rules are created, then they have to be detected and removed as soon as possible. Preserving only the rules that proved to be useful avoids the number of rules in the controller growing above a reasonable limit.

Since we create new rules while there is a significant error in the value prediction, if necessary, we could end up generating complete rules (provided we do not limit the number of rules in our controller). In this case, and assuming that the more specific the rule the more accurate the value prediction, our system would behave as a normal table-based reinforcement-learning algorithm: Only the most specific rules (i.e., the most relevant ones) would be used to evaluate actions and, as explained before, the statistics for these rules would be exactly the same as those in table-based reinforcement-learning algorithms. Thus, in the limit, our system can deal with the same type of problems as non-generalizing reinforcement-learning algorithms. However, we regard this limit situation as very improbable and we impose limits to the number of rules in our controllers. Observe that this asymptotic convergence to a table-based reinforcement learning is only possible because we use a winner-takes-all strategy in the action evaluation. With a weighted-sum strategy, the value estimation for the non-complete rules possibly present in the controller would be added to that of complete rules leading to an action evaluation different from that of table-based reinforcement-learning algorithms.

## 5. The Partial Rule Approach in Context

The categorizability assumption is closely related with complexity theory principles such as the Minimum Description Length (MDL) that has been used by authors such as Schmidhuber (2002) to bias learning algorithms. All these complexity results try to formalize the well-known "Occam's Razor" principle that enforces choosing the simplest model from a set of otherwise equivalent models.

Boutilier, Dean, and Hanks (1999) presents a good review on representation methods to reduce the computational complexity of planning algorithms by exploiting the particular characteristics of a given environment. The representation based on partial rules can be seen as another of these representation systems. However, the partial rule is just a representation formalism that, without the bias introduced by the categorizability assumption, would not be efficient enough to be applied to realistic applications.

The partial-rule formalism can be seen as a generalization of that of the XCS classifier systems described by Wilson (1995). This XCS learning system aims at determining a set of *classifiers* (that are combinations of features with an associated action) with their associated value and relevance predictions. The main difference between this approach and ours is that Wilson's work pursues a generic learner and we bias the learning process using the categorizability assumption. This allows us to use an incremental rule-generation strategy that is likely to be more efficient in robotic problems. Additionally, the categorizability assumption also modifies the way in which the value for a given action is evaluated: Wilson's approach uses a weighted sum of the predictions of the classifier advocating for each action to determine the expected effect of that action, while, to fulfill with the categorizability assumption (i.e., to minimize the number of feature detectors and elementary actions involved in a given evaluation), we propose to use a winner-takes-all strategy. This is a critical point since the winner-takes-all strategy takes full advantage of the categorizability assumption and because it allows the partial-rule system to asymptotically converge to a table-based reinforcement-learning system. This is not the case when a weighted sum strategy is used. Furthermore, in the XCS formalism there is no generalization in the action space and, as already commented, this is a requirement in robotic-like applications.

In general, reinforcement learning does not pay attention to the necessity of generalizing in the space of actions, although some exceptions exists. For instance, the work of Maes and Brooks (1990) includes the possible execution of elementary actions in parallel. However this system does not include any mechanism detecting interactions between actions and, thus, the coordination of actions relies on sensory conditions. For instance, this system has difficulties detecting that the execution of two actions results always (i.e., independently of the active/inactive feature detectors) in positive/negative reward.

The CASCADE algorithm by Kaelbling (1993) learns each bit of a complex action separately. This algorithm presents a clear sequential structure where the learning of a given action bit depends on all the previously learned ones. In our approach there is not a predefined order in the learning of the outputs and the result is a more flexible learning schema.

In multiagent learning (Claus & Boutilier, 1998; Sen, 1994; Tan, 1997) the objective is to learn an optimal behavior for a group of agents trying to cooperatively solve a given task. Thus, in this field, as in our case, multiple actions issued in parallel have to be considered. However, one of the main issues in multiagent learning, the coordination between the different learners is irrelevant in our case since we only have one learner.

Finally, the way in which we define complex actions from elementary actions has some points in common with the works in reinforcement learning where macro-actions are defined as the learner confronts different tasks (Sutton, Precup, & Singh, 1999; Drummond, 2002). However, the useful combinations of elementary actions detected by our algorithm are only guaranteed to be relevant for the task at hand (although they are likely to be also relevant for related tasks).

## 6. Experiments

We show the results of applying our learning algorithm to two robotics-like simulated problems: robot landmark-based navigation and legged robot walking. The first problem is simpler (although it includes more delayed reward) and we use it to clearly describe the workings of the algorithm. The second problem approaches a realistic robotic application, our objective in the long term. We use the two examples to compare the performance of our learning system with that of generalizing and non-generalizing reinforcement-learning algorithms. The confronted problems are different enough to show the generality of the proposed learning system.

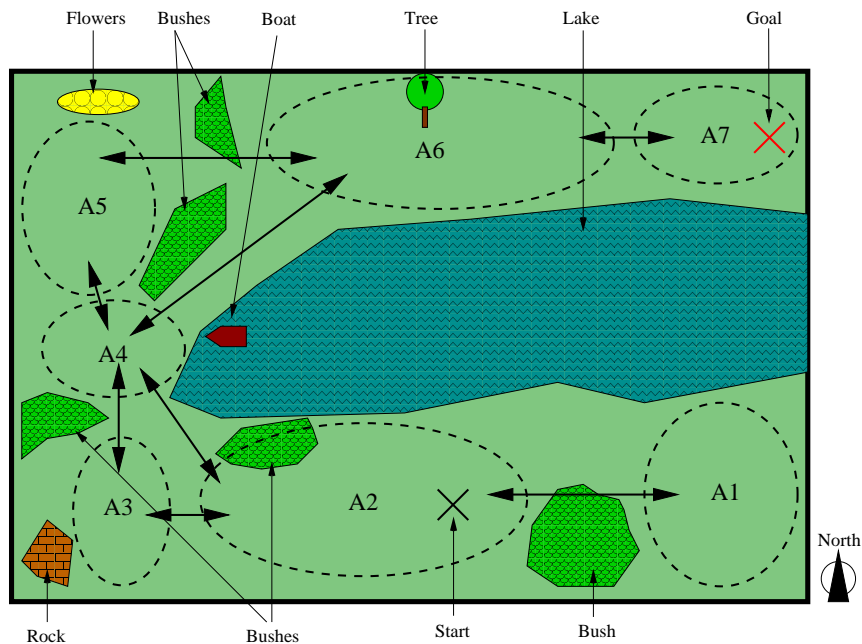

Figure 1: Landscape for the simple landmark-based navigation task. The landscape is divided in areas (the dashed ovals) where the same subsets of landmarks are visible.

## 6.1 Simulated Landmark-Based Navigation

We confront a simple simulated landmark-based navigation task in the forest-like environment shown in Figure 1. The objective for the learner is to go from the start position (marked with a cross at the bottom of the figure) to the goal position where there is the food (marked with a cross at the top right corner of the environment). The agent can neither walk into the lake nor escape from the depicted terrain.

The agent can make use of some binary landmark (i.e., feature) detectors to identify its position in the environment and to decide which action to execute next. In the example, the landmark detectors of the agent are:

1. **Rock detector:** Active when the rock is seen.

2. **Boat detector:** Active when the boat is seen.

3. **Flower detector:** Active when the bunch of flowers is seen.

4. **Tree detector:** Active when the tree is seen.

5. **Bush detector:** Active whenever a bush is seen.

6. **Water detector:** Active when there is water nearby.

7. **Bird detector:** Active when there is a bird flying over the agent.

8. **Cow detector:** Active when there is a cow nearby.

9. **Sun detector:** Active when the sun is shining.

10. **Cloud detector:** Active when it is cloudy.

Of these detectors, only the first 5 are relevant for the task. The *water detector* is always active, and the rest of landmark detectors become active at random. These 10 landmark detectors can differentiate between $2^{10} = 1024$ situations.

We simplify the problem by clustering the possible positions of the learner in the environment in 7 areas (shown in Figure 1): each area includes the positions from which the same set of relevant landmarks can be seen.

As far as actions is concerned, we use three actions for the West-East movement of the robot: move to the West (denoted as $W$), stay in the same place ($-$), move to the East ($E$). The other three indicate movement along the North-South dimension (move to the North $N$, stay in the same latitude $-$, move to the South $S$). These two independent groups of three actions can be combined giving rise to 9 different actions (move North-West, North, North-East, etc.). We assume that when the agent executes one of these actions, it does not stop until the nearest area of the terrain in the direction of the movement is reached. When the agent tries to move into the lake or out of the terrain, it remains in the same position it was. Figure 1 shows all the possible transitions between contiguous areas in the environment.

With the just described landmark detectors and elementary actions the maximum possible order of a given rule is 12, and we can define up to 944784 ($3^{10}4^2$) syntactically different partial rules. Only taking into account all the rules with one feature detector and one elementary action (that are the ones initially included in the controller) we have 90 different partial rules.

The agent only receives reward (with value 100) when it reaches the goal. Consequently, this is a problem with delayed reward since the agent must transmit the information provided by the reward signal to those actions and situations not directly related with the observation of reward.

The parameters of the partial-rule learning algorithm we used for this task were $\gamma = 0.9$, $\beta = 0.99$, $\eta = 5$, $\alpha = 0.1$, $\tau = 5$, $\mu = 200$ and, $\lambda = 0.95$ (see Appendix A for a detailed description of the parameters). Observe that, with a maximum number of partial rules $\mu = 200$ and an initial controller containing 90 rules, little room is left for the generation of rules with order higher that 2.

The learning is organized in a sequence of trials. Each trial consists in placing the learner in the starting position and letting it move until the goal is reached, allowing the execution of at most 150 actions to reach the goal. When performing optimally, only three actions are required to reach the objective from the starting position.

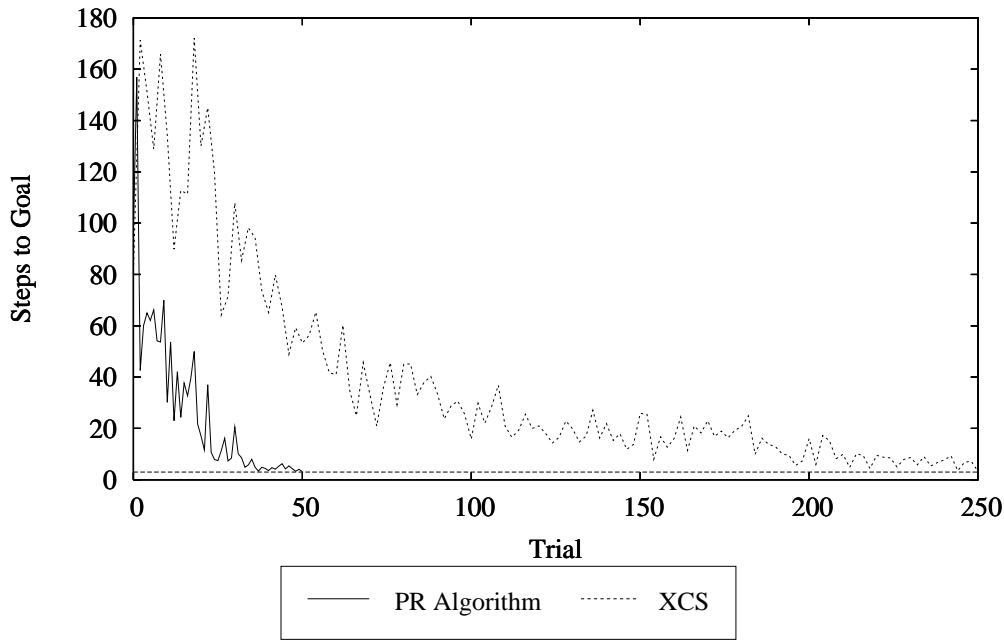

Figure 2: Performance on the landmark-based navigation task. Results shown are the average over 10 runs.

Figure 2 shows that, after 40 learning trials, the agent approaches the optimal behavior (represented by the flat dashed line at $y = 3$).

The dashed line in Figure 2 is the performance of a XCS in this problem. To perform this test, we used the implementation of Wilson's XCS developed by Butz (1999). To make XCS work in the same search space as the partial-rule algorithm, we modified the XCS implementation to be able to deal with non-binary actions. No other modification, but parameter adjustment, were introduced in the original code. The results presented here corresponds to the average of 10 runs using the set of parameters that gave a better result. Nominally, these parameters were: learning rate $\alpha = 0.1$, decay rate $\gamma = 0.9$, maximum number of classifiers $\mu = 200$ (however, the initial set is empty), the genetic algorithm is applied in average every 5 time steps, the deletion experience is 5, the subsume experience is 15, the fall off rate is 0.1, the minimum error 0.01, a prediction threshold of 0.5, the crossover probability is 0.8, the mutation probability 0.04 and the initial *don't care* probability 1/3. The prediction and the fitness of new classifiers are initialized to 10 and the error to 0. A detailed explanation of the meaning of those parameters is provided by Wilson (1995) and also by the comments in the code of Butz (1999).

We can see that the XCS reaches the same performance of the partial-rule approach, but using about four times more trials. This difference in performance is partially explained by XCS's lack of generalization in the action space. However this factor is not that relevant in this case since the action space has only two dimensions. The main factor that explains the better performance of the partial-rule approach is the bias introduced by the categorizability

93

| $t$ | Position | $V$ | Action | Winner Rule | $q_w$ | $e_w$ | Guess |
|---|---|---|---|---|---|---|---|
| 1 | A2 | 81 | $(W, N)$ | $w_1 = (v(Rock, \neg Boat), c(W, N))$ | 80.63 | 1.16 | 79.87 |
| | | | $(W, -)$ | $w_2 = (v(Rock, Water), c(W))$ | 79.73 | 2.19 | 77.65 |
| 2 | A4 | 90 | $(-, N)$ | $w_3 = (v(Boat, Tree), c(N))$ | 89.61 | 2.04 | 88.88 |
| | | | $(E, N)$ | $w_4 = (v(Tree), c(E, N))$ | 90.0 | 0.0 | 89.86 |
| | | | $(E, -)$ | $w_5 = (v(\neg Rock, Boat), c(E))$ | 86.71 | 4.58 | 79.56 |
| 3 | A6 | 100 | $(E, -)$ | $w_6 = (v(\neg Bush), c(E, -))$ | 100.0 | 0.0 | 99.87 |

Table 1: Partial execution trace for the landmark-based navigation task. Elementary action $-$ means no movement along the corresponding dimension. At each time step $t$, the action with the highest guess is executed. After time step 3, the goal is reached.

assumption that is not present in the XCS system and that, in this case, allows a more efficient learning process. XCS in more powerful than our partial-rule approach in the sense that XCS makes no assumption about the categorizability of the environment, while we assume it as high. The result of the XCS learning process includes the identification of the degree of categorizability of the environment and in our case this is, in some sense, pre-defined. The generality of the XCS, however, produces a slower learning process.

If we initialize the classifiers in a XCS with a high *don't care* probability and we initialize the rules in the partial-rule algorithm so that no generalization is used in the action space (i.e., if all rules include a command for each motor), then the two systems become closer. In this case, the main (but not the only) difference between the two approaches is the assumption on the relation between the inputs and the value: While XCS assumes a linear relation, we assume the environment to be categorizable, or, what is the same, we assume the value to depend on only few of the inputs. Due to this difference, when confronted to the same problem, the two systems would learn the *same* policy and the *same* values for each action, but the values would be computed using different rules with different values, and this is so independently of the parameter/rule initialization used in each case. The system with a smaller learning time would be that with an assumption closer to the reality. The results obtained in the particular example presented above show that the categorizability assumption is more valid and our hypothesis is that this would be the case in most robotics-like applications.

Table 1 shows the evaluation of some actions in the different situations the agent encounters on its path from the start to the goal after 50 learning trials. Analyzing this trace, we can extract some insight about how the partial-rule learning algorithm works.

For instance, at time step 1, we see that rule $w_2 = (v(Rock, Water), c(W))$ is used to determine the value of action $(W, -)$. Since the landmark detector *Water* is always active, this rule is equivalent to $w = (v(Rock), c(W))$, which is one of the rules used to generate $w_2$. If we examine the statistics for $w$ we find that $q_w = 74.70$ and $e_w = 15.02$. Obviously, the value distributions of $q_w$ and $q_{w_2}$ look different (74.70 vs. 79.73 and 15.02 vs. 2.19). This is because $w_2$ has been generated on later stages of the learning and, thus, its statistics have been updated using a subsample of the values used to adjusts the statistics of $w$. In this particular case, $q_w$ has been updated 250 times while $q_{w_2}$ has been updated only 27 times. As learning continues, both distributions will become more similar and rule $w_2$ will be eventually eliminated.

In Table 1, we can see that sometimes there are some non-optimal actions that get an evaluation close to the optimal ones. For this reason, the agent executes, some times, non-optimal actions and this increases the number of steps necessary to reach the goal. In general, the adjustment of the statistics of the rules can solve this problem but, in this particular case, we need to create new rules to fix the situation. For instance, at time step 2, when the value for rule $w_4$ is increased towards 90, the value of rules active at this time step and proposing actions in accordance with the action of rule $w_4$ also converge toward 90. So, in the long term, a rule proposing just action $(N)$ can get a value close to 90. In the absence of more specific rules, this rule can be used to estimate the value of an action such as $(-, N)$ and, due to the probabilistic nature of the action selection procedure, this action can, eventually, be executed delaying the agent from reaching the goal by 1 time step. However, the execution of $(-, N)$ results in an error in the value prediction and, thus, in the creation of new rules to better characterize the situation. As soon as a specific rule for action $(-, N)$ is generated, the error is no longer repeated.

At time step 3, we see that rule $w_6 = (v(\neg Bush), c(E, -))$ has a value of 100 with error 0 but the guess for this rule is 99.87. This is because the maximum confidence ($\beta$) is lower than 1.0 (0.99 in this case) and this makes the agent to keep always a certain degree of exploration.

If the agent only receives reward when the task is totally achieved, the function value for each situation can be computed as $V(s) = \gamma^{n-1} r$ with $n$ the distance (in actions) from situation $s$ to the target one and $r$ the reward finally obtained. In Table 1, we can see that situations get the correct evaluation: $80.63 (\sim 81 = 100 \cdot 0.9^2)$ for $A2$, $90 (= 100 \cdot 0.9)$ for $A4$, and 100 for $A6$.

Observe that this problem can be solved using only 200 partial rules out of the 9216 possible situation-action combinations in this domain. So, we can say that the problem is certainly categorizable. The main conclusion we can extract from this toy example is that, in a particular case in which the confronted problem was categorizable, the presented algorithm has been able to determine the relevant rules and to adjust their values (including the effect of the delayed reward) so that the optimal action can be determined for each situation.

## 6.2 Gait Generation for a Six-Legged Robot

We also applied our algorithm to the task of learning to generate an appropriate gait (i.e., the sequence of steps) for a six-legged robot (Figure 3). To apply the learning algorithm to the real robot would be possible, but dangerous: in the initial phases of the learning the robot would fall down many times damaging the motors. For this reason we used a simulator during the learning and, afterward, we applied the learned policy to the real robot.

The problem of learning to walk with a six legged robot has been chosen by many authors before as a paradigmatic robotic-learning problem. For instance, Maes and Brooks (1990) implemented a specific method based on immediate reward to derive the preconditions for each leg to perform the step. Pendrith and Ryan (1996) used a simplified version of the six-legged walking problem to test an algorithm able to deal with Non-Markovian spaces of states and Kirchner (1998) presented a hierarchical version of Q-learning to learn the low-level movements of each leg, as well as a coordination scheme between the low-level learned behaviors. Ilg, Mühlfriedel, and Berns (1997) introduced a learning architecture based on self-organizing neural networks, and Kodjabachia and Meyer (1998) proposed an evolutionary strategy to develop a neural network to control the gait of the robot. Vallejo and Ramos (2000) used a parallel genetic algorithm architecture and Parker (2000) described an evolutionary computation where the robot executes the best controller found up to a given moment while a new optimal controller is computed in an off-line simulation. All these algorithms are usually tested on flat terrain with the aim of generating periodic gaits (i.e., gaits where the sequence of steps is repeated cyclically). However, for general locomotion (turns, irregular terrain, etc) the problem of free gait generation needs to be considered.

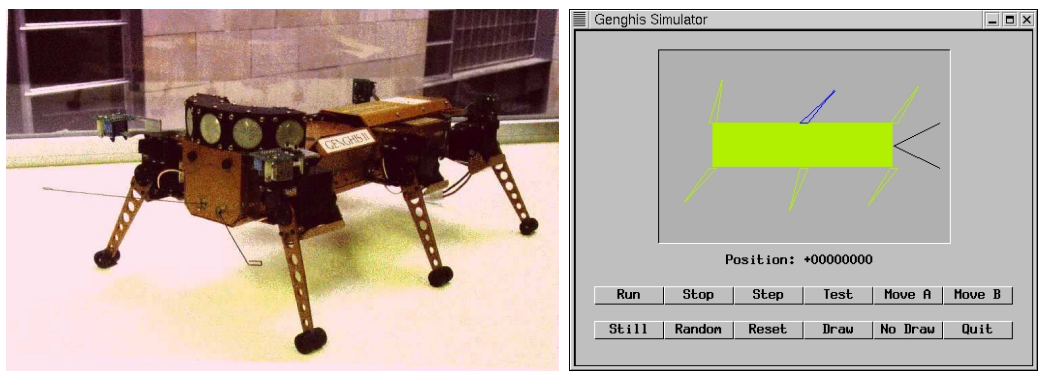

Figure 3: The Genghis II walking robot and its 2D simulation environment.

Our simulator (see Figure 3) allows the controller to command each leg of the robot in two independent degrees of freedom (horizontal and vertical) and it is able to detect when the robot is in an unstable position (in our robot this happens when two neighboring legs are in the air simultaneously). Using this simulator, we implemented the behaviors described by Celaya and Porta (1996) except those in charge of the gait generation. Therefore, the task to be learned consists in deciding at every moment which legs must step (that is, leave the ground and move to an advanced position), and which must descend or stay on the ground to support and propel the body.

We defined a set of 12 feature detectors that, due to our experience on legged robots, we knew could be useful in different situations for the gait-generation task:

- **In_the_air(x):** Active if the leg $x$ is in the air.

- **Advanced(x):** Active if leg $x$ is more advanced than its neighboring leg in a clockwise circuit around the robot.

Attending to the activation and non-activation of these 12 feature detectors, we can differentiate between 4096 different situations.

On the action side, we work with two different elementary actions per leg: one that issues the step of the leg and another that descends the leg until it touches the ground.

Thus, the cardinality of the set of elementary actions is 12 and, at each time step, the robot issues an action containing 6 elementary elements (one per leg). Thus, we can think of each leg as a virtual motor that accepts two possible values, 0 to remain in contact with the ground and 1 to perform the step.

The reward signal includes two aspects:

- **Stability:** If an action causes the robot to fall down, a reward of $-50$ is given.

- **Efficiency:** When the robot does not fall down, a reward equal to the distance advanced by the robot is given. Observe that when legs descend to recover contact with the ground no advance of the robot is obtained but this movement is necessary to be able to get reward in next time steps. So, we have again a problem with delayed reward.

The most efficient stable gait is the *tripod gait* in which two sets of three non-adjacent legs step alternately. Using this gait, the robot would obtain a reward of 0 (when one group of three legs are lifted and advanced) followed by a reward of 50 (when the legs in contact with the ground move backward as a reaction to the advance of legs moved in the previous time step). Thus, the optimal average reward is 25.

In the experiments, the robot is set in an initial posture with all the legs in contact with the ground but in a random advance position.

Figure 4 shows results of applying the partial-rule algorithm compared with those obtained using standard Q-learning with 4096 distinct states and 64 different actions.

For the partial-rule algorithm, we used the following set of parameters: $\gamma = 0.2$, $\beta = 0.99$, $\eta = 22$, $\alpha = 0.1$, $\tau = 150$, $\mu = 10000$ and, $\lambda = 0.95$ (see Appendix A for a description of these parameters). For Q-learning, the learning rate is set to $\alpha = 0.5$ and we use an action selection rule that performs exploratory actions with probability 0.1.

In Figure 4, we can see that the stability subproblem (i.e., not falling down, which corresponds to getting a reward greater than zero) is learned very quickly. This is because, in the stability subproblem, we can take advantage of the generalization provided by using separate elementary actions and, with a single rule, we can avoid executing several dangerous actions. However, the advance subproblem (i.e., getting a reward close to 25) is learned slowly. This is because little generalization is possible and the learning system must generate very specific rules. In other words, this sub-problem is less categorizable than the stability one.

As in the landmark-based navigation example discussed in the previous section, we observe that the controller contains some (slightly) overly general rules that are responsible for the non optimal performance of the robot. However, we don't regard this as a problem since we are more interested in efficiently learning a *correct enough* policy for the most frequent situations than in finding optimal behaviors for all particular cases.

Figure 5 shows the performance of Q-learning over a longer run using different exploration rates. This shows that Q-learning can eventually converge to an optimal policy but with many more iterations than our approach (about a factor of 10). Observe that a lower exploration rate allows the algorithm to achieve higher performance (around 19 with a learning rate of 0.1 and around 24 with learning rate 0.01) but using a longer period. With a careful adjustment of the exploration rate we can combine an initial *faster* learning with a better convergence in the long term. Experiments with Q-learning using learning rates other than 0.5 showed insignificant differences compared to the results shown here.

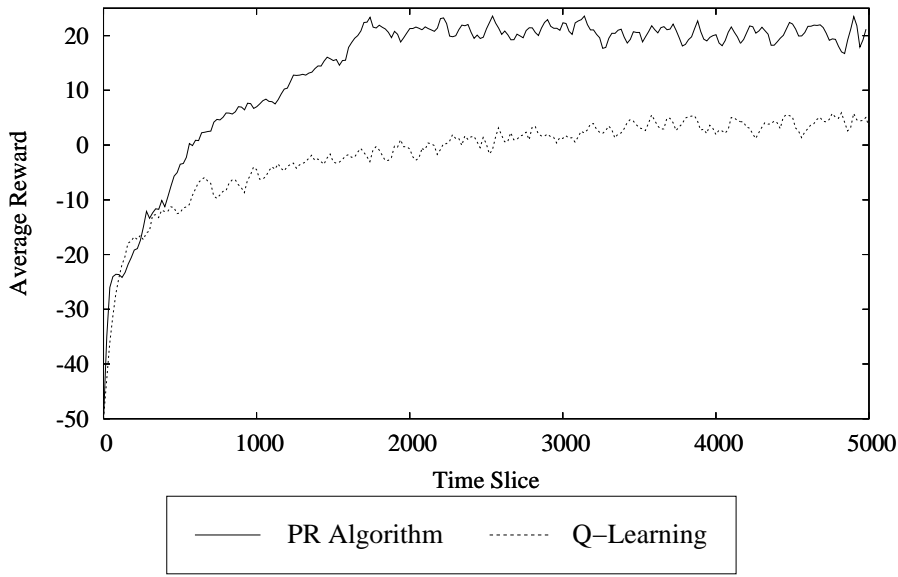

Figure 4: Performance of the partial-rule approach compared with standard Q-learning. Results are the smoothed average of 10 experiments.

The advantage of our algorithm over non-generalizing ones is increased in problems in which some of the sensors provide information not related to the task. To test this point, we set up an experiment in which 6 feature detectors that become active randomly were added to the 12 initial ones. With these new features, the number of possible combinations of feature activations increases, and so does the number of states considered by Q-learning. Figure 6 shows the comparison between our algorithm and Q-learning for this problem. Q-learning is not able to learn a reasonable gait strategy in the 5000 time steps shown in the figure, while the performance of the partial-rule algorithm is almost the same as before. This means that the partial-rule algorithm is able to detect which sets of features are relevant and use them effectively to determine the robot's behavior. It is remarkable that, in this case, the ratio of memory used by our algorithm with respect to that used by non-generalizing algorithms is below 0.2%. This exemplifies how the performance of the non-generalizing algorithms degrades as the number of features increases, while this is not necessarily the case using the partial-rule approach.

The importance of the generation of partial rules in the improvement of the categorization can be seen comparing the results obtained for the same problem with and without this mechanism (Figure 7). The results show that the task cannot be learned using only partial rules of order 2. The only aspect of the gait-generation problem that can be learned with rules of order 2 is to avoid lifting a leg if one of its neighboring legs is already in the

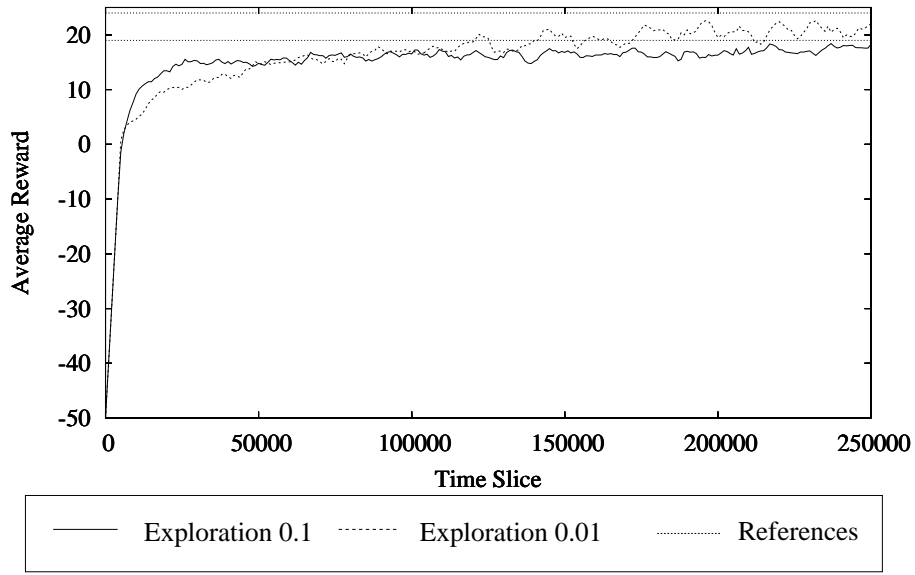

Figure 5: Performance of the Q-learning algorithm with different exploration rates. The reference values 19 and 24 are the upper bound of the performance attainable when using exploration rate 0.1 and 0.01.

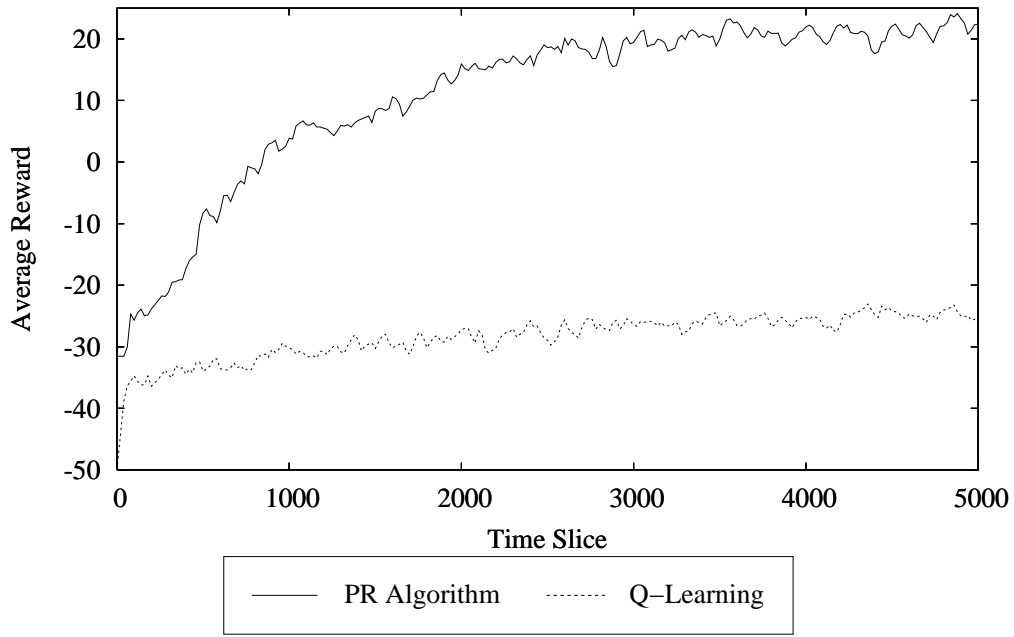

Figure 6: Performance of our algorithm compared with Q-learning when there are irrelevant features.

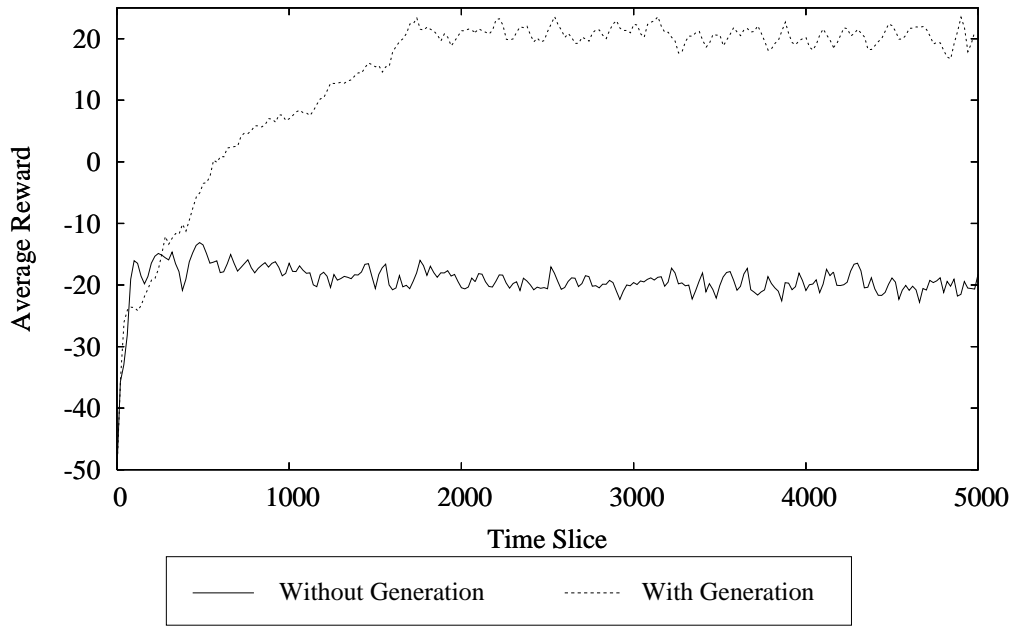

Figure 7: Performance with and without the partial-rule generation procedure.

air. For instance, a rule such as

$$v(\mathbf{In\_the\_air}(1)) \rightarrow c(\mathbf{Step}(2)),$$

forecasts a highly relevant negative reward and this prevents leg 2 from being raised when leg 1 is in the air.

Rules with order higher than 2 (i.e., not provided to the robot in the initial controller) are necessary, for instance, to avoid raising two neighboring legs simultaneously. A rule like

$$v(\neg\mathbf{In\_the\_air}(1)) \rightarrow c(\mathbf{Step}(1), \mathbf{Step}(2))$$

becomes active when the robot evaluates any action that implies raising leg 1 and leg 2 at the same time. Since the value prediction of this rule is very negative and its relevance is high, the action under evaluation would be discarded, preventing the robot from falling down. Similar rules have to be generated for each pair of neighboring legs. To make the robot advance, we need to generate rules with even higher order.

In Figure 8, we can see the performance of the algorithm when we start the learning process from a correct rule set (i.e., a rule set learned in a previous experiment), but with all the statistics initialized to 0. In this experiment, we can compare the complexity of learning only the values for the rules compared with the complexity of learning the rules and their value at the same time. We can see that when only the values for the rules need to be learned the learning process is about two times faster than in the normal application of the algorithm.

In a final experiment, we issue frequent changes in the heading direction of the robot (generated randomly every 10 time steps). In this way, periodic gaits become suboptimal and the controller should produce a free gait, i.e., a gait that includes a sequence of steps without any periodic repetition.

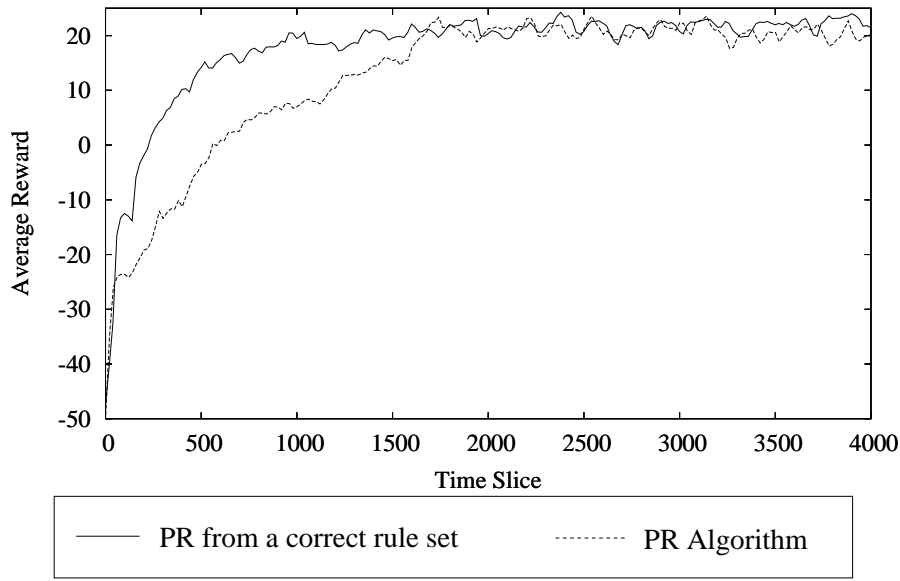

Figure 8: Performance of the partial-rule approach when learning is started from a correct rule set compared with the standard approach where rules are also learned.

In this case, we focus on the advance subproblem and, thus, we introduced some hand-crafted rules to the initial controller to prevent the robot from falling down. These rules are of the form:

**if** *leg i is lifted* **then** *execution of action a results in value −50 with confidence 1,*

where $a$ is any of the actions that lift one of the two legs that are contiguous to $i$.

The set of parameters we used in this case was: $\gamma = 0.2$, $\beta = 0.99$, $\eta = 5$, $\alpha = 0.1$, $\tau = 150$, $\mu = 10000$ and, $\lambda = 0.95$.

Figure 9 shows the average results obtained using the partial-rule learning algorithm compared with those obtained with our best hand-coded gait-generation strategy. In the figure, the horizontal dashed line shows the average performance using the best gait-generation strategy we have implemented (Celaya & Porta, 1998). It can be seen that the learned gait-generation strategy (the increasing continuous line) produces a performance similar to that of our best hand-coded strategy and that, in some cases, it even outperforms it. Figure 10 shows a situation where a learned controller produces a better behavior than our hand coded one. Using the hand-coded strategy, the robot starts to walk raising two legs (3 and 6) and, in few time steps it reaches a state from which the tripod gait is generated. Initially, leg 2 is more advanced than legs 1 and 4 and, in general, it is suboptimal to execute a step with a leg when its neighboring legs are less advances that itself. In this particular case however, this general rule does not hold. The learned strategy detects this exception and generates the tripod gait from the very beginning resulting in a larger advance of the robot in the initial stages of the movement.

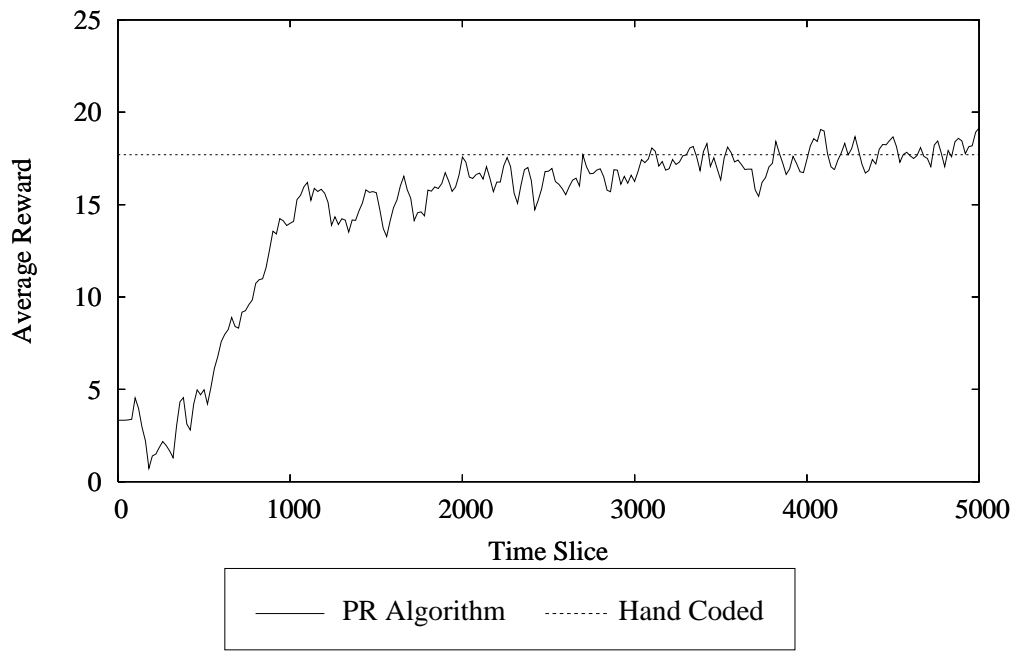

Figure 9: Performance of the partial-rule approach when learning a free gait.

## 7. Conclusions

In this paper, we have introduced the *categorizability assumption* that states that a robot can be driven to achieve a given task using only simple rules: i.e., rules including a reduced set of feature detectors and elementary actions. This assumption is supported by our experience within the behavior-based approach where controllers are formed by sets of rules with relatively simple conditions and actions. We have shown that a learning algorithm based on the categorizability assumption allows a large speed up in the learning process in many realistic robotic applications with respect to existing algorithms.

To exploit the categorizability assumption both in the observations and action spaces, we have introduced a new representation formalism based on the concept of *partial rules* and not on the concepts of independent states and independent actions that are the kernel of many existing reinforcement-learning approaches.

The introduction of the *partial-rule* concept provides a large flexibility on how problems are formalized. With the same structure and algorithms, we can confront problems with generalization in the perception side (usually considered in reinforcement learning), in the action side (usually not considered), or in both of them.

When no generalization is possible at all via partial rules, we have to use *complete rules*: rules involving all the available inputs and outputs. In this case, the partial-rule approach is equivalent to the non-generalizing reinforcement learning. The algorithm we have presented can, if necessary, generate *complete rules* and, consequently, it can, in principle, solve any problem that can be solved using a traditional reinforcement-learning algorithm. However, we take the categorizability assumption as valid and so, the generation of complete rules is an extreme case that is only likely to occur in a very limit situation. Therefore, in our approach, we forego generality in order to increase the efficiently of the learning process in the class of problems we want to address.

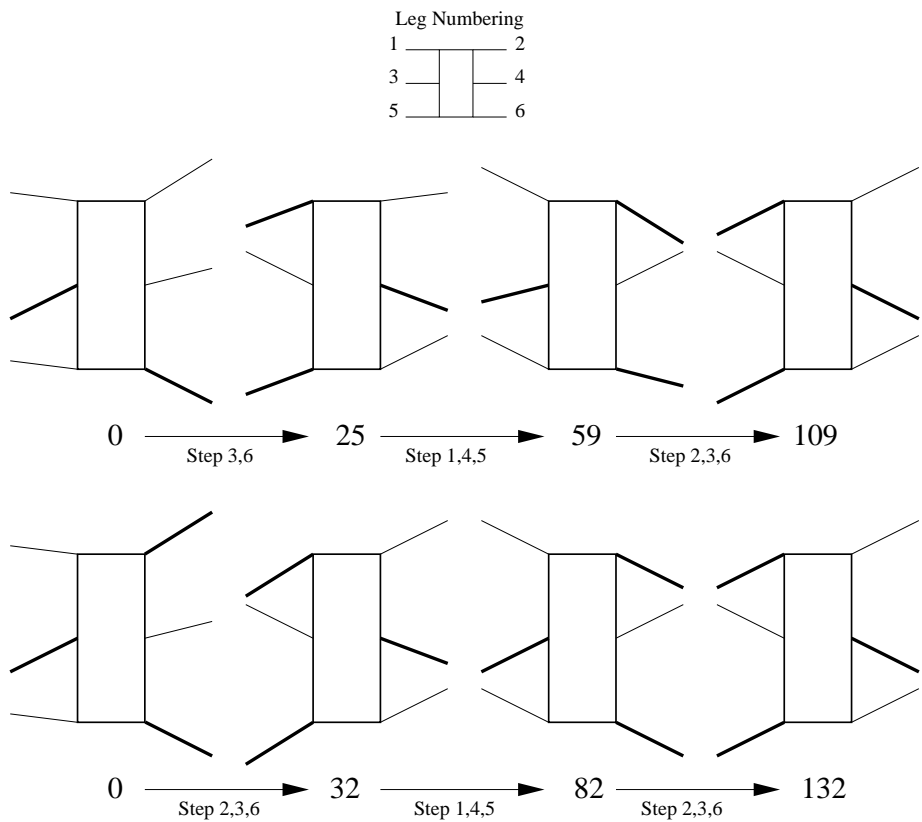

Figure 10: The hand-programmed gait strategy (top sequence) vs. a learned one (bottom sequence). The advance position of the robot at each snapshot is indicated below each picture.

Another advantage of the partial-rule framework is that it allows the easy and robust introduction of initial knowledge in the learning process in the form of rules that are easily understood by the programmer. This is in contrast with usual reinforcement-learning algorithms where the introduction of initial knowledge is, in general, rather difficult.

In the partial-rule approach, there is a subtle change of emphasis as to the main goal of learning: While in most work in reinforcement learning the emphasis is on learning the value of each action in each state, our main purpose is to learn the relevance of (subsets of) elementary actions and feature detectors. If the relevant subsets of elementary actions and feature detectors are identified, the learning becomes straightforward.

The main limitation of our work is that it is not possible to know a priori (except for trivial cases) whether or not an environment is categorizable by a given robot. Non-generalizing reinforcement learning implicitly assumes that the environment is non-categorizable and that, consequently, all the possible combination of features and actions have to be taken into account separately. Our approach assumes just the opposite: that the environment is categorizable and, so, only reduced combinations of features and actions need to be taken into account. The drawback of using the non-generalizing approach is that robotic tasks become intractable because of the *curse of dimensionality*. With generalization techniques this problem can be partially alleviated, but not enough in general. In our approach we take a more radical approach in order to be much less affected by the curse of dimensionality: we introduce a strong bias in the learning process to drastically limit the use of combinations of features and actions.

We have tested the partial-rule learning algorithm in many robotic-inspired problems and two of them have been discussed in this paper (landmark based-navigation and six-legged robot gait generation) and the categorizability assumption proved to be valid in all cases we tested. The algorithm out-performs generalizing and non-generalizing reinforcement-learning algorithms in both memory requirements and convergence time. Additionally, we have shown that our approach scales well when the number of inputs increases, while the performance of existing algorithms is largely degraded. This is a very important result that lets us think that it could be possible to use our approach to control more complex robots, while the use of existing approaches has to be discarded.

From the work presented in this paper, we can extract two main proposals. First, to apply reinforcement learning to agents with many sensors and actuators, we should concentrate our efforts in determining the relevance of inputs and outputs and, second, to achieve efficient learning in complex environments it could be necessary to introduce additional assumptions into the reinforcement-learning algorithms, even at the risk of losing generality.

## Acknowledgments

The authors would like to express their gratitude to the anonymous reviewers of the paper. Their contributions toward improving the quality of this paper are relevant enough to be considered, in some sense, as a co-authors of the paper. The shortcomings still in the paper can only be attributed to the nominal authors.

The second author has been partially supported by the Spanish *Ministerio de Ciencia y Tecnología* and FEDER funds, under the project DPI2003-05193-C02-01 of the *Plan Nacional de I+D+I*.

## Appendix A: The Partial-Rule Learning Algorithm

In this appendix, we describe in detail the approach described in the main body of the paper.

```
Partial Rule Learning Algorithm
    (Initialize)
        FD ← Set of features detectors
        EA ← Set of elementary actions
        C ← {w_∅} ∪ {(v(fd), c(ea)), (v(¬fd), c(ea))|fd ∈ FD, ea ∈ EA}
        For each w in C
            q_w ← 0
            e_w ← 0
            i_w ← 0
        endfor
        ē ← 0
    Do for each episode
        C' ← {w ∈ C|w is active}
        Repeat (for each step in the episode):
            (Action Selection)
                Action Evaluation          (Computes guess(a') ∀a')
                a ← arg max {guess(a')}
                        ∀a'
                Execute a
            (System Update)
                r_a ← Reward generated by a
                C'_ant ← C'
                C' ← {w ∈ C|w is active}
                Statistics Update
                Partial-Rule Management
        until terminal situation
    enddo
```

Figure 11: The partial-rule learning algorithm. Text inside parentheses are comments. The *Action Evaluation*, *Statistics Update*, and *Partial-Rule Management* procedures are described next.

The partial-rule learning algorithm (whose top level form is shown in Figure 11) stores the following information for each partial rule

- the value (i.e., the discounted cumulative reward) estimation $q_w$,

- the error estimation $e_w$, and

- the confidence index $i_w$.

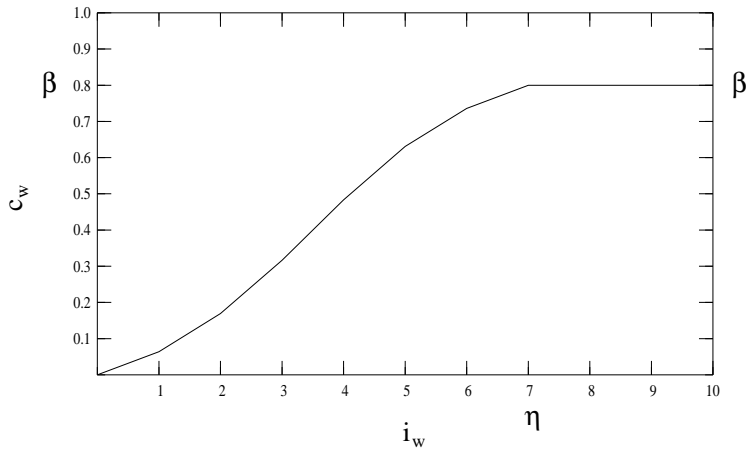

Figure 12: Confidence function with $\eta$=7 and $\beta$=0.8.

To estimate the confidence on $q_w$ and $e_w$ we use a confidence index $i_w$ that, roughly speaking, keeps track of the number of times the partial rule is used. The confidence is derived from $i_w$ using a *confidence_function* in the following way:

$$c_w = confidence\_function(i_w),$$

where the *confidence_function* is a non-decreasing function in the range $[0, \beta]$. $\beta$ should be less than 1 since, in this way, the system always keeps a certain degree of exploration and, consequently, it is able to adapt to changes in the environment. Different confidence schemes can be implemented by changing the *confidence_function*. In our implementation, we use a sigmoid-like function (see Figure 12) that increases slowly for low values of $i_w$ reducing the confidence provided by the first obtained rewards. In this way we avoid a premature increase of the confidence (and, thus, a decrease in the error and in the exploration) for insufficiently-sampled rules. A parameter ($\eta$) determines the point at which this function reaches the top value $\beta$.

Additionally, the confidence index is used to define the learning rate (i.e., the weight of new observed rewards in the statistics update). For this purpose we implement a MAM function (Venturini, 1994) for each rule:

$$m_w = \max\{\alpha, 1/(i_w + 1)\}.$$

Using a MAM-based updating rule, we have that, the lower the confidence, the higher the effect of the last observed rewards on the statistics, and the faster the adaptation of the statistics. This adaptive learning rate strategy is related to those presented by Sutton (1991) and by Kaelbling (1993), and contrasts with traditional reinforcement-learning algorithms where a constant learning rate is used.

After the initialization phase, the algorithm enters in a continuous loop for each task episode consisting in estimating the possible effects of all actions, executing the most promis-ing one, and updating the system so that its performance improves in the future. The system update includes the statistics update and the partial-rule management.

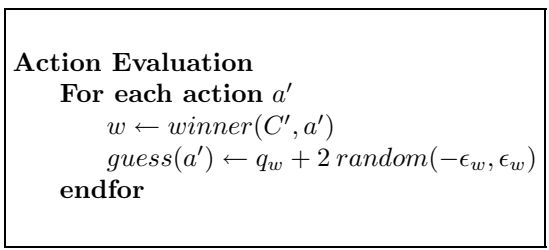

Figure 13: Action Evaluation procedure.

## Action Evaluation

The simplest procedure to get the estimated value for actions is a *brute-force* approach consisting of the independent evaluation of each one of them. In simple cases, this approach would be enough but, when the number of valid combinations of elementary actions (i.e., of actions) is large, the separate evaluation of each action would take long time, increasing the time of each robot decision and decreasing the reactivity of the control. To avoid this, Appendix B presents a more efficient procedure to get the value of any action.

Figure 13 summarizes the action-evaluation procedure using partial rules. The value for each action is *guessed* using the most relevant rule for this action (i.e., the *winner* rule). This winner rule is computed as

$$winner\,(C', a) = \arg\max_{\forall w \in C'(a)} \{\rho_w\},$$

where $\rho_w$ is the relevance of rule $w$

$$\rho_w = \frac{1}{1 + \epsilon_w}.$$

The value estimation using the winner rule is selected at random (uniformly) from the interval

$$I_w = [q_w - 2\epsilon_w, q_w + 2\epsilon_w],$$

with

$$\epsilon_w = e_w\,c_w + \overline{e}\,(1 - c_w).$$

Here, $\overline{e}$ is the average error on the value prediction (i.e., the value error prediction of the empty rule, $w_\emptyset$).

## Statistics Update

In the statistics-update procedure (Figure 14), $q_w$ and $e_w$ are adjusted for all rules that were active in the previous time step and proposed a partial command in accordance with $a$ (the last executed action).

**Statistics Update**
    **if** terminal situation **then**
        $v \leftarrow 0$
    **else**
        $v \leftarrow \max_{\forall a'}\{q_w | w = winner(C', a')\}$
    **endif**
    $q \leftarrow r_a + \gamma \, v$
    **For each** $w = (v, c)$ **in** $C'_{ant}$
        **if** $c$ is in accordance with $a$ **then**
            **if** $q \in I_w$ **then**
                $i_w \leftarrow i_w + 1$
            **else**
                $i_w \leftarrow \min(\eta - 1, i_w - 1)$
            **endif**
            $q_w \leftarrow q_w \, (1 - m_w) + q \, m_w$
            $e_w \leftarrow e_w \, (1 - m_w) + |q_w - q| \, m_w$
        **endif**
    **endfor**
    $\overline{e} \leftarrow e_{w_\emptyset}$

Figure 14: Statistics update procedure.

Both $q_w$ and $e_w$ are updated using a learning rate ($m_w$) computed using the MAM function, which initially is 1, and consequently, the initial values of $q_w$ and $e_w$ have no influence on the future values of these variables. These initial values become relevant when using a constant learning rate, as many existing reinforcement-learning algorithms do.

If the observed effects of the last executed action agree with the current estimated interval for the value ($I_w$), then the confidence index is increased by one unit. Otherwise, the confidence index is decreased allowing a faster adaptation of the statistics to the last obtained, surprising values of reward.

**Partial-Rule Management**

This procedure (Figure 15) includes the generation of new partial rules and the removal of previously generated ones that proved to be useless.

In our implementation, we apply a heuristic that produces the generation of new partial rules when the value prediction error exceeds $\overline{e}$. In this way, we concentrate our efforts to improve the categorization on those situations with larger errors in the value prediction.

Every time a wrong prediction is made, at most $\tau$ new partial rules are generated by combination of pairs of rules included in the set $C'_{ant}(a)$. Recall that this set includes the rules active in the previous time step and in accordance with the executed action $a$. Thus, these are the rules related with the situation-action whose value prediction we need to improve.

The combination of two partial rules $w_1 \oplus w_2$ consists of a new partial rule with a partial view that includes all the features included in the partial views of either $w_1$ or $w_2$ and with a partial command that includes all the elementary actions of the partial commands of either $w_1$ or $w_2$. In other words, the feature set of $w_1 \oplus w_2$ is the union of the feature sets in $w_1$ and in $w_2$ and the elementary actions in $w_1 \oplus w_2$ are the union of those in $w_1$ and those in $w_2$. Note that, since both $w_1$ and $w_2$ are in $C'_{ant}(a)$, they have been simultaneously active and they are in accordance with the same action and, thus, they can not be incompatible (i.e., they can not include inconsistent features or elementary actions).

In the partial-rule creation, we bias our system to favor the combination of those rules $(w_i)$ whose value prediction $(q_{w_i})$ is closer to the observed one $(q)$. Finally, the generation of rules lexicographically equivalent to already existing ones is not allowed.

According to the categorizability assumption, only low-order partial rules are required to achieve the task at hand. For this reason, to improve efficiency, we limit the number of partial rules to a maximum of $\mu$. However, our partial-rule generation procedure is always generating new rules (concentrating on those situations with larger error). Therefore, when we need to create new rules and there is no room for them, we must eliminate the less useful partial rules.

A partial rule can be removed if its value prediction is too similar to some other rule in the same situations.

The similarity between two rules can be measured using the normalized degree of intersection between their value distributions and the number of times both rules are used simultaneously:

$$similarity(w, w') = \frac{\|I_w \cap I_{w'}\|}{\max\{\|I_w\|, \|I_{w'}\|\}} \frac{U(w \oplus w')}{\min\{U(w), U(w')\}},$$

where $U(w)$ indicates the number of times rule $w$ is actually used.

The similarity assessment for any pair of partial rules in the controller is too expensive and, in general, determining the similarity of each rule with respect to those from which it was generated (that are the rules we tried to refine when the new rule was created) is sufficient. Thus, based on the above similarity measure, we define the *redundancy* of a partial rule $w = (w_1 \oplus w_2)$ as:

$$redundancy(w) = \max\{similarity(w, w_1), similarity(w, w_2)\}.$$

Observe that with $w = (w_1 \oplus w_2)$, we have that $w \oplus w_1 = w$ and $U(w) \leq U(w_1)$. Therefore

$$\frac{U(w \oplus w_1)}{\min\{U(w), U(w_1)\}} = \frac{U(w)}{\min\{U(w), U(w_1)\}} = \frac{U(w)}{U(w)} = 1.$$

The same reasoning can be done with $w_2$ and, consequently,

$$redundancy(w) = \max\{\frac{\|I_w \cap I_{w_1}\|}{\max\{\|I_w\|, \|I_{w_1}\|\}}, \frac{\|I_w \cap I_{w_2}\|}{\max\{\|I_w\|, \|I_{w_2}\|\}}\}.$$

When we need to create new rules but the maximum number of rules $(\mu)$ has been reached, the partial rules with a redundancy above a given threshold $(\lambda)$ are eliminated. Since the redundancy of a partial rule can only be estimated after observing it a number of

---

**Partial Rule Management**

$w \leftarrow winner(C'_{ant}, a)$

**if** $|q_w - q| > \overline{e}$       *(If it is time to create new rules)*

   *(Partial Rule Elimination)*

      *(Test if there is no room for new rules)*

      **if** $\|C\| > \mu - \tau$ **then**

         *(Rule elimination based on redundancy)*

         $C \leftarrow C - \{w \in C \mid redundancy(w) > \lambda\}$

         *(Rule elimination based on creation error)*

         **if** $\|C\| > \mu - \tau$ **then** *(If there is still no room)*

            $SC \leftarrow$ The $\tau$ partial rules from $C$ with:
               - Lowest *creation_error(w)*, and
               - *creation_error(w)* $< |q_w - q|$

            $C \leftarrow C - SC$

         **endif**

      **endif**

   *(Partial Rule Generation)*

      $t \leftarrow 0$

      **while** $\|C\| < \mu$ **and** $t < \tau$

         *(Create a new rule $w'$)*

         Select two different rules $w_1$, $w_2$ from $C'_{ant}(a)$
            preferring those that minimize
               $|q_{w_i} - q| \, c_{w_i} + \overline{e} \, (1 - c_{w_i})$

         $w' = (w_1 \oplus w_2)$

         $creation\_error(w') \leftarrow |q_w - q|$

         *(Insert the new rule in the controller)*

         $C \leftarrow C \cup \{w'\}$

         $t \leftarrow t + 1$

      **endwhile**

**endif**

---

Figure 15: Partial Rule Management procedure. The value of $q$ is calculated in the Statistics Update procedure and $a$ is the last executed action.

times, the redundancy of the partial rules with low confidence indexes is set to 0, so that they are not immediately removed after creation.

Observe that, to compute the redundancy of a rule $w$, we use the partial rules from which $w$ was derived. For this reason, a rule $w'$ cannot be removed from a controller $C$ if there exists any rule $w \in C$ such that $w = w' \oplus w''$. Additionally, in this way we eliminate first the useless rules with higher order.

## Appendix B: Efficient Action Evaluation

In non-generalizing reinforcement learning the cost of executing a single learning step can be neglected. However, algorithms with generalization in the spaces of sensors and/or actuators are not so simple and the execution time of each iteration can be increased substantially. In an extreme case, this increase can limit the reactivity of the learner and this is very dangerous when working with an autonomous robot.

The most expensive procedure of our algorithm is that of computing the value of all actions (i.e., all valid combinations of elementary actions). The cost of this procedure is especially critical since it is used twice in each step: once to get the guess of each action (in the *Action Evaluation* procedure detailed in Figure 13) and again to get the goodness of the new achieved situation after the action execution (when computing the $v$ value in the *Statistics Update* procedure detailed in Figure 14). A trivial re-order of the algorithm can avoid the double use of this expensive procedure at each learning step: we can select the action to be executed next at the same time that we evaluate the goodness of the new achieved situation. The drawback of this re-order is that the action is selected without taking into account the information provided by the *last* reward value (the goodness of the situation is assessed before the value adjustment). However, this is not a problem in tasks that require many learning steps.

Even if we use the action-evaluation procedure only once per learning step, we have to optimize it as much as possible since the brute-force approach described before, which evaluates each action sequentially, is only feasible for simple problems.

The action-evaluation method presented next is based on the observation that many of the actions would have the same value since the highest relevant partial rule at a given moment would provide the value to all actions that are in accordance with the partial command of the rule. The separate computation of the value of two actions that would end up evaluated using the same rule is a waste of time. This can be avoided by performing the action evaluation attending to the set of active rules in the first place and not to the set of possible actions, as the brute-force approach does.

Figure 16 shows a general form of the algorithm we propose. In this algorithm, partial rules are considered one at a time, ordered from the most relevant rule to the least relevant one. The partial command of the rule under consideration ($co_w$) is used to process all the actions that are in accordance with that partial command. This already processed sub-set of actions need not to be considered any more in the action-evaluation procedure. While the rules are processed, we update the current situation assessment ($v$) and the action to be executed next ($a$) attending, respectively, to the value prediction ($q_w$) and the guess ($g_w$) of the rules.

Observe that partial rules can be maintained sorted by relevance by the statistics update procedure, since it is in this procedure where rule relevance is modified. When the relevance of a rule is changed, its position in the list can be also modified accordingly. In this way we do not have to re-sort the list of rules every time we want to apply the procedure just described.

When elementary actions are of the form ($m \leftarrow k$) with $m$ a motor and $k$ a value in the range of possible values for that motor, the above algorithm can be implemented in an especially efficient way since there is no need to explicitly compute the set of actions $A$.

```
Action Evaluation
    (Initialization)
        L ← List of active rules sorted by relevance.
        EA ← Set of elementary actions
        A ← Set of combinations of EA
        v ← −∞        (Situation assessment)
        a ← ∅         (Optimal action)
        g ← −∞        (Optimal action value prediction)
    (Process)
        w ← first_element(L)
        do
            co_w ← partial command of w
            g_w ← q_w + 2 random(−ε_w, ε_w)
            A_w ← {a ∈ A|co_w in accordance with a}
            if q_w > v then
                v ← q_w
            endif
            if  g_w > g then
                g ← g_w
                a ← co_w
            endif
            A ← A − A_w
            w ← next_element(L)
        until A ≠ ∅
```

Figure 16: General form of the proposed situation-assessment and action-selection procedure.

In this case (see Figure 17 and 18), we construct a decision tree using motors as decision attributes and that groups in the same leaf all those actions evaluated by the same partial rule (all actions removed from the set $A$ in each iteration of the algorithm in Figure 16).

Each internal node of the tree classifies the action according to one of the motor commands included in the action. These internal nodes store the following information:

- Partial command: A partial command that is in accordance with all the action classified under the node. This partial command can be constructed by collecting all the motors whose values are fixed in the nodes from the root of the tree to the node under consideration.

- Motor: The motor used in this node to classify actions. When a node is *open* (i.e., we have still not decided to which motor to attend) the motor value is set to a ⊥. A node can be *closed* by deciding which motor to pay attention to (and adding the corresponding subtrees) or by converting the node into a leaf.

```
Action Evaluation
   (Initialization)
      L ← List of active rules sorted by relevance.
      v ← −∞
      a ← ∅
      g ← −∞
      tree ← new_node(∅_c)
      open ← 1
      closed ← 0
   (Process)
      w ← first_element(L)
      do
         g_w ← q_w + 2 random(−ε_w, ε_w)
         Include Rule(tree, w, g_w)
         w ← next_element(L)
      until closed = open
```

Figure 17: Top level algorithm of the efficient action evaluation algorithm. At the end of the algorithm, $v$ is the goodness of the current situation to be used in the *Statistics Update* algorithm (see Figure 14), $a$ is the action to be executed next and *guess* its expected value. The "Include Rule" procedure is detailed in next figure.

- Subtrees: This is a list of the subtrees that start in that node. Each subtree has an associated value that corresponds to one of the possible actions executable by the motor of the node. All the actions included in a given subtree have an elementary action such as $(m \leftarrow k)$ where $m$ is the motor of the node and $k$ is the value corresponding to this subtree.

The leaves of the tree have information about the value of the actions classified in that leaf. This information is represented with the following set of attributes for each leaf:

- Value: The expected value for all the actions classified in this leaf. The maximum of this value for all leaves is used to assess the goodness, $v$, of a new achieved situation.

- Guess: The value altered with noise for exploratory reasons. The leaf with a maximal guess is the set of actions from where to select the action to be executed next.

- Relevance: The relevance of the value predictions (of both the value and the guess).

- Partial command: A partial command that is in accordance with all the actions classified in that leaf. As in the case of internal nodes, this partial command can be constructed by collecting all the motors whose values are fixed from the root of the tree to the leaf under consideration.

---

**Include rule**$(n, w, g_w)$
  **if not**$(is\_leaf(n))$ **then**
      $co_w \leftarrow command(w)$
      $co_n \leftarrow command(n)$
      **if** $motor(n) \neq \perp$ **then**     *(Closed Node: Search for compatible sub-nodes)*
          **if** $\exists ea \in co_w$ **with** $motor(ea) = motor(n)$ **then**
              Include Rule$(get\_subtree(value(ea), n), w, g_w)$
          **else**
              **for all** $s$ **in** $subtrees(n)$ **do**
                  Include Rule$(s, w, g_w)$
              **endfor**
          **endif**
      **else**     *(Open Node: Specialize the node)*
          **if** $co_w - co_n \neq \varnothing$ **then**     *(Extend a node)*
              $ea \leftarrow action\_in(co_w - co_n)$
              $set\_motor(n, motor(ea))$
              $closed \leftarrow closed + 1$
              **for all** $k$ **in** $values(motor(ea))$ **do**
                  $new\_subtree(n, \{k, new\_node(co_n \oplus (motor(ea) \leftarrow k))\})$
                  $open \leftarrow open + 1$
              **endfor**
              Include Rule$(n, w, g_w)$
          **else**     *(Transform a node into a leaf)*
              $transform\_to\_leaf(n, q_w, g_w, \rho_w, co_w)$
              $closed \leftarrow closed + 1$
              **if** $q_w > v$ **then**
                  $v \leftarrow q_w$
              **endif**
              **if** $g_w > guess$ **then**
                  $g \leftarrow g_w$
                  $a \leftarrow co_w$
              **endif**
          **endif**
      **endif**
  **endif**

---

Figure 18: The *Include rule* algorithm searches for nodes from node $n$ with a partial command compatible with the partial command of rule $w$ and extends those nodes to insert a leave in the tree.

At a given moment, the inclusion of a new partial rule in the tree produces the specialization of all open nodes compatible with the rule (see Figure 18). We say that an open node $n$ is compatible with a given rule $w$ if the partial command of the node $co_n$ and the partial command of the rule $co_w$ does not assign different values to the same motor. The specialization of an open node can result in the extension of the node (i.e., new branches are added to the tree under that node) or in the transformation of this node into a leaf. A node is extended when the partial command of the rule affects some motors not included in the partial command of the node. This means that there are some motor values not taken into account in the tree but that have to be used in the action evaluation according to the rule under consideration. When a node is extended, one of the motors not present in the above layers of the tree is used to generate a layer of open nodes in the current node. After that, the node is considered as closed and the inclusion rule procedure is repeated for this node (with different effects because now the node is closed). When all the motors affected by the partial command of the rule are also affected by the partial command of the node, then the node is transformed into a leaf storing the value, guess, and relevance attributes extracted from the information associated with the rule.

| Partial rules | | $q$ | $e$ | $\rho$ | $guess$ |
|---|---|---|---|---|---|
| Partial View | Partial Command | | | | |
| $TRUE_v$ | $(m_1 \leftarrow v_1) \wedge (m_2 \leftarrow v_1)$ | 5 | 0.1 | 0.83 | 5.1 |
| $TRUE_v$ | $(m_1 \leftarrow v_1)$ | 7 | 0.9 | 0.52 | 6.5 |
| $TRUE_v$ | $(m_2 \leftarrow v_1) \wedge (m_3 \leftarrow v_1)$ | 8 | 2.0 | 0.33 | 6.0 |
| $TRUE_v$ | $(m_2 \leftarrow v_1)$ | 3 | 3.1 | 0.24 | 6.2 |
| $TRUE_v$ | $(m_1 \leftarrow v_0)$ | 2 | 3.5 | 0.22 | 5.3 |
| $TRUE_v$ | $(m_2 \leftarrow v_0)$ | 10 | 3.6 | 0.21 | 4.1 |
| $TRUE_v$ | $(m_3 \leftarrow v_1)$ | 1 | 4.0 | 0.20 | 5.2 |
| $TRUE_v$ | $(m_3 \leftarrow v_0)$ | 6 | 4.5 | 0.18 | 12.7 |

Table 2: Set of rules of the controller. The values $q$ and $e$ are stored and the $\rho$ and $guess$ are computed from them. We define all partial views as $TRUE_v$ to indicate that they are active in the current time step.

The process is stopped as soon as we detect that all nodes have been closed (i.e. all the external nodes of the tree are leaves). In this case, the rules still to be processed can have no effect in the tree form and, consequently, are not useful for action evaluation. If a rule is consistently not used for action evaluation, it can be removed from the controller.

A toy-size example can illustrate this tree-based action-evaluation algorithm. Suppose that we have a robot with three motors that accept two different values (named $v_0$ and $v_1$). This produces a set of 8 different actions. Suppose that, at a given moment, the robot controller includes the set of rules shown in Table 2. In the *Action Evaluation* algorithm (Figure 17), rules are processed from the most to the least relevant one expanding an initially empty tree using algorithm in Figure 18. The inclusion of a rule in the tree results in an extension of the tree (see stages B, D and E in Figure 19) or in closing branches by converting open nodes into leaves (stages C and F). In this particular case the tree becomes completely closed after processing 5 rules out of the 8 active rules in the controller. At the end of the process, we have a tree with five leaves. Three of them include two actions and the other two only represent a single action. Using the tree we can say that the value of the situation in which the tree is constructed, $v$, is 8 (this is given by the leaf circled with a solid line in the figure). Additionally, the next action to be executed is of the form $(m_1 \leftarrow v_1, m_2 \leftarrow v_0, m_3 \leftarrow \sharp)$ where '$\sharp$' represents any possible action. This optimal action is given by the leaf circled with a dashed line that is the leaf with a larger *guess* value.

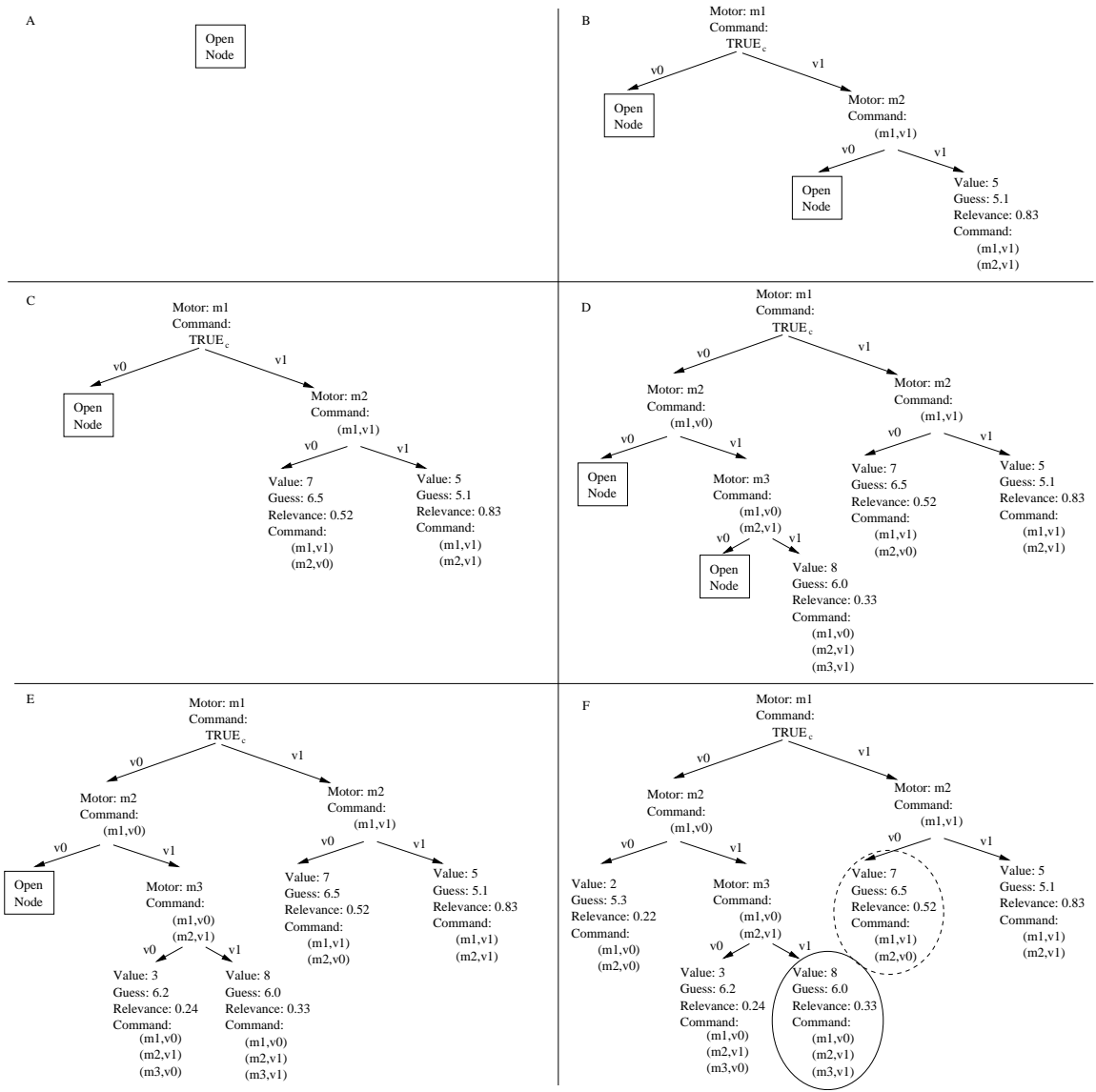

Figure 19: Six different stages during the construction of the tree for action evaluation. Each stage corresponds to the insertion of one rule from Table 2.

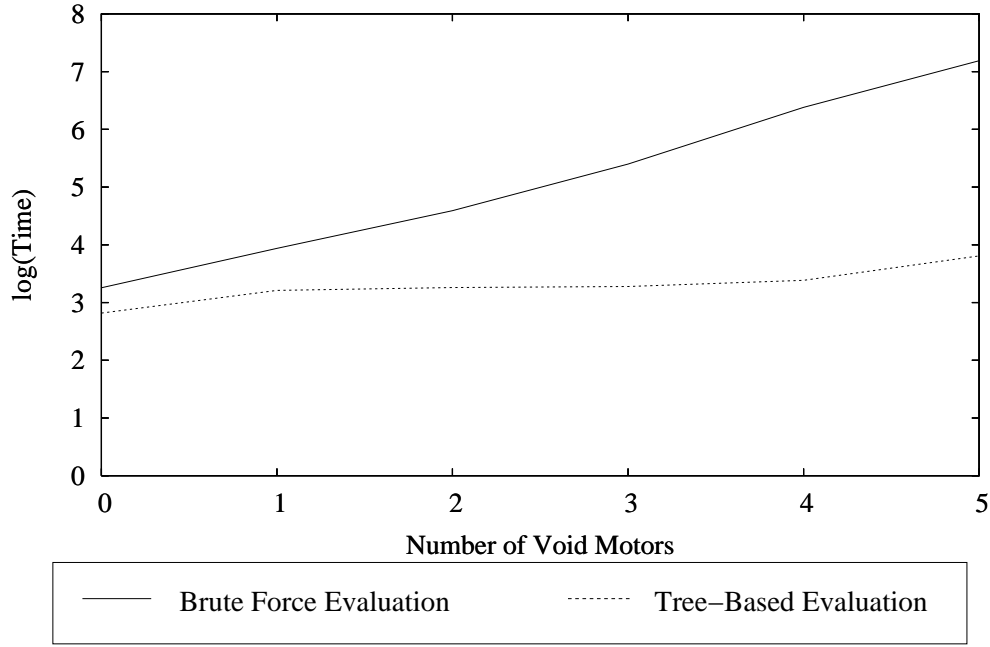

Figure 20: Log of the execution time (in seconds) of the brute-force approach vs. the tree-based one.

The cost of our algorithm largely depends on the specific set of partial rules to be processed. In the worst case, the cost of our algorithm is:

$$O(n_r \, l^{n_m}),$$

with $n_r$ the number of rules, $n_m$ the number of motors and, $l$ the maximal range of values accepted by the motors. This is because, in the worst case, to insert a given rule, we have to visit all the nodes of a maximally expanded tree (i.e., a tree where each node has $l$ subtrees and where all the final nodes of the branches are still opened). The number of nodes of such a tree is

$$\sum_{i=0}^{n_m} l^i = \frac{l^{n_m+1}-1}{l-1} = O(l^{n_m}).$$

We can transform the cost expression taking into account that $l^{n_m}$ is the total number of possible combinations of elementary actions ($n_c$) or, in other words, the total amount of actions. Therefore, the cost of the presented algorithm is

$$O(n_r \, n_c).$$

On the other hand, the cost of the brute-force approach is always

$$\Theta(n_r \, n_c).$$

So, in the worst case, the cost of the presented algorithm is of the same order as the cost of the brute-force approach. However, since at most $l$ rules would be enough to close a maximally expanded tree (one rule for the different values of the motor used in the last still-open layer of the tree), the cost of the tree-based algorithm would be, on average, much smaller than that of the brute-force approach.

Figure 20 exemplifies the different performance of the brute-force action-evaluation procedure and the tree-based one. The figure shows the time taken in the execution of the toy example of Section 6.1. For this experiment, we defined some *void motors* or motors whose actions have no effect in the environment. As it can be seen, as the number of void motors increases, the cost of the tree-based evaluation is significantly less than that of the brute-force approach.

## Appendix C: Notation

Uppercase are used for sets, and Greek letters represent parameters of the algorithms.

| | |
|---|---|
| $S$ | Set of states. |
| $s, s'$ | Individual states. Full views. |
| $n_s$ | Number of states. |
| $FD = \{fd_i \mid i = 1..n_f\}$ | Set of feature detectors. |
| $v(fd_{i_1}, \ldots, fd_{i_k})$ | Partial view of order $k$. |
| $A$ | Set of actions of the robot. |
| $n_a$ | Number of actions. |
| $EA = \{ea_i \mid i = 1..n_e\}$ | Set of elementary actions. |
| $n_m$ | Number of motors of the robot. |
| $ea_i = (m_i \leftarrow k)$ | Elementary action that assigns value $k$ to motor $m_i$. |
| $c(ea_{i_1}, \ldots, ea_{i_k})$ | Partial command of order $k$. |
| $a = (ea_1, \ldots, ea_{n_m})$ | Action. Combination of elementary actions. Full command. |
| $w = (v, c)$ | Partial rule composed by partial view $v$ and partial command $c$. |
| $w_\emptyset$ | The empty partial rule. |
| $w_1 \oplus w_2$ | Composition of two partial rules. |
| $C = \{w_i \mid i = 1..n_r\}$ | Controller or set of partial rules. |
| $\mu$ | Maximum number of elements of $C$. |
| $C', C'_{ant}$ | Subset of rules active at a given time step and at the previous one. |
| $C'(a)$ | Active rules with a partial command in accordance with $a$. |
| $q_w$ | Expected value of the partial rule $w$. |
| $e_w$ | Expected error in the value estimation of the partial rule $w$. |
| $\overline{e}$ | Average error in the value prediction. |
| $i_w$ | Confidence index. |
| $c_w$ | Confidence on the statistics of the partial rule $w$. |
| $\beta$ | Top value of the confidence. |
| $\eta$ | Index where the confidence function reaches the value $\beta$. |
| $\epsilon_w = e_w \, c_w + \overline{e} \, (1 - c_w)$ | Error in the return prediction of the partial rule $w$. |
| $\rho_w = 1/(1 + \epsilon_w)$ | Relevance of rule $w$. |
| $I_w = [q_w \pm 2\epsilon_w]$ | Value interval of the partial rule $w$. |
| $m_w$ | Updating ratio for the statistics of the partial rule $w$. |
| $\alpha$ | Learning rate. Top value of $m_w$. |
| $U(w)$ | Number of times rule $w$ has been used. |
| $winner(C', a)$ | Most relevant active partial rule w.r.t. action $a$. |
| $guess(a)$ | Most reliable value estimation for action $a$. |
| $r_a$ | Reward received after the execution of $a$. |
| $\gamma$ | Discount factor. |
| $v$ | Goodness of a given situation. |
| $q = r_a + \gamma v$ | Value of executing action $a$ in given situation. |
| $\tau$ | Number of new partial rules created at a time. |
| $\lambda$ | Redundancy threshold used for partial-rule elimination. |